%% file: usenix.tex
\newcommand{\attname}{CachedAttention\xspace}

\newcommand{\storagename}{AttentionStore\xspace}

\documentclass[letterpaper,twocolumn,10pt]{article}

\usepackage{titlesec}

\usepackage{siunitx}
\usepackage{url}
\usepackage{usenix}
\usepackage{epsfig}
\usepackage{booktabs}
\usepackage{xspace}
\usepackage{multirow}
\usepackage{listings}
\usepackage{amsmath}
\usepackage{subcaption}
\usepackage{caption}
\usepackage{filecontents}
\usepackage{xurl}
\usepackage{ulem}
\usepackage{wrapfig}
\usepackage{pifont}
\usepackage{comment}
\usepackage{xcolor}

\usepackage[misc]{ifsym}

\definecolor{codegreen}{rgb}{0,0.6,0}
\definecolor{codegray}{rgb}{0.5,0.5,0.5}
\definecolor{codepurple}{rgb}{0.58,0,0.82}
\definecolor{backcolour}{rgb}{0.95,0.95,0.92}
\lstdefinestyle{mystyle}{
    backgroundcolor=\color{backcolour},   
    commentstyle=\color{codegreen},
    keywordstyle=\color{magenta},
    numberstyle=\tiny\color{codegray},
    stringstyle=\color{codepurple},
    basicstyle=\ttfamily\footnotesize,
    breakatwhitespace=false,         
    breaklines=true,                 
    captionpos=b,                    
    keepspaces=true,                 
    numbers=left,                    
    numbersep=5pt,                  
    showspaces=false,                
    showstringspaces=false,
    showtabs=false,                  
    tabsize=2
}
\usepackage[linesnumbered, ruled]{algorithm2e}

\SetKwFunction{FunctionName}{FunctionName}
\SetKwProg{Fn}{Function}{:}{}
\SetKwRepeat{Do}{do}{while}%
\usepackage[affil-it]{authblk}
\makeatletter
\renewcommand\AB@affilsepx{\quad\quad \protect\Affilfont}
\def\thanks#1{\protected@xdef\@thanks{\@thanks
        \protect\footnotetext{#1}}}
\makeatother

\begin{document}

\date{}

\title{\Large \bf Cost-Efficient Large Language Model Serving for Multi-turn Conversations with \attname}

\author[1,*\thanks{*Work done during their internship at Huawei Cloud.}]{Bin Gao}
\author[2,*]{Zhuomin He}
\author[1]{Puru Sharma}
\author[1]{Qingxuan Kang}
\author[1]{Djordje Jevdjic}
\author[3]{Junbo Deng}
\author[3]{\\Xingkun Yang}
\author[3]{Zhou Yu}
\author[3,$\dagger$ \thanks{$\dagger$Corresponding author: Pengfei Zuo (pengfei.zuo@huawei.com).}]{Pengfei Zuo}

\affil[1]{National University of Singapore}
\affil[2]{Shanghai Jiaotong University}
\affil[3]{Huawei Cloud}
\maketitle


\subsection*{Abstract}
Interacting with humans through multi-turn conversations is a fundamental feature of large language models (LLMs). However, existing LLM serving engines executing multi-turn conversations are inefficient due to the need to repeatedly compute the key-value (KV) caches of historical tokens, incurring high serving costs.
To address the problem, this paper proposes CachedAttention, a new attention mechanism that enables reuse of KV caches across multi-turn conversations, significantly reducing the repetitive computation overheads. CachedAttention maintains a hierarchical KV caching system that leverages cost-effective memory/storage mediums to save KV caches for all requests. To reduce KV cache access overheads from slow mediums, CachedAttention employs layer-wise pre-loading and asynchronous saving schemes to overlap the KV cache access with the GPU computation. To ensure that the KV caches to be accessed are placed in the fastest hierarchy, CachedAttention employs scheduler-aware fetching and eviction schemes to consciously place the KV caches in different layers based on the hints from the inference job scheduler. To avoid the invalidation of the saved KV caches incurred by context window overflow, CachedAttention enables the saved KV caches to remain valid via decoupling the positional encoding and effectively truncating the KV caches.
Extensive experimental results demonstrate that CachedAttention significantly decreases the time to the first token (TTFT) by up to 87\%, improves the prompt prefilling throughput by up to 7.8$\times$ for multi-turn conversations, and reduces the end-to-end inference cost by up to 70\%. 

\input{tex/introduction}
\input{tex/background}

\input{tex/design}

\input{tex/evaluation}
\input{tex/related}

\input{tex/conclusion}

\section*{Acknowledgments}
This work was partly supported by Huawei Cloud, the Advanced Research and Technology Innovation Centre (ARTIC) at the National University of Singapore under grant FCT-RP1 A-0008129-00-00, and the Ministry of Education in Singapore grants A-0008143-00-00 and A-0008024-00-00.

\bibliography{sample}
\bibliographystyle{plain}

\end{document}

%% file: tex/introduction.tex
\section{Introduction}
With impressive performance on a wide variety of tasks, large language models (LLMs) have ushered in a new era of generative applications~\cite{Chatgpt, touvron2023llama, team2023gemini}. However, serving these generative applications with LLMs is very expensive due to the LLM inference employing a large number of GPUs. Given the high demand for generative applications, reducing the cost of inference becomes crucial. 

Engaging in multi-turn conversations with humans is an essential capability of LLMs~\cite{wu2023autogen, wang2023mint}. These multi-turn conversations help LLMs comprehend context, user intent, and emotional nuances, enhancing their ability to respond appropriately. Based on the ShareGPT data~\cite{sharegpt_sharegptraw_2024}, a widely-used real dataset collected from ChatGPT, 73\% of conversations involve multiple turns, as analyzed in Section~\ref{sec:background-multi-turn}. 

However, executing multi-turn conversations in current LLM serving engines is highly inefficient, as it requires a large number of repetitive computations, incurring high serving costs. During a single turn of conversation, the LLM engine stores intermediate data, key-value (KV) pairs~\cite{aminabadi_deepspeed_2022, pope_efficiently_2023, kwon_efficient_2023}, in the limited high-bandwidth memory (HBM) on GPUs. When that conversation ends and the conversation session becomes inactive, the LLM engine generally discards the KV cache associated with that session, to free up space in the HBM for other active sessions. When the session becomes active again, i.e., the user sends the next message in the conversation, the LLM engine computes the whole KV cache again. This leads to repetitive computation of the same KV cache, wasting valuable GPU computation resources.
With the number of conversation turns increases, the repetitive computation overhead linearly increases. Our analysis based on ShareGPT shows that up to 99\% of the prefilling cost comes from repetitive computation for the KV cache, as presented in Section~\ref{sec:background-multi-turn}.

To reduce the serving cost and improve the inference performance, this paper proposes \textit{\attname}, a new attention mechanism
that enables the reuse of KV caches across multi-turn conversations rather than discarding them. When a conversation session becomes inactive, \attname saves the corresponding KV cache in a KV caching system, named \textit{\storagename}. Upon the resumption of the same session, \attname loads and reuses the saved KV cache from \storagename, thereby eliminating the overhead of the repetitive computation. However, building such an efficient KV caching system for multi-turn conversations presents significant challenges.

Firstly, \storagename serves as the external storage for GPUs and is attached to the GPUs via low-speed links. The use of \storagename brings about significant access overhead due to the need to transfer KV caches between HBMs and \storagename. The access overhead of KV caches is in the critical path of inference execution. This is because GPUs can only perform the computation of an inference job after successfully loading its corresponding KV cache into HBMs. Likewise, the subsequent inference jobs need to wait until the KV caches from the previous jobs are moved out of the HBMs if the HBM space is not enough. 
To reduce the KV cache loading overheads, \attname uses a layer-wise pre-loading scheme to overlap the time of loading the KV cache with the inference computation layer by layer. To reduce the KV cache saving overheads, \attname develops an asynchronous saving scheme that overlaps the time of saving KV caches with the inference computation.

Secondly, the KV caches occupy a large amount of storage space that continuously expands during conversations. Prior works have attempted to reduce the inefficiency of repetitive KV computation by retaining the KV caches across multi-turn conversations in HBMs~\cite{lmdeploy_2023, zheng2023efficiently}. However, this quickly exhausts the limited HBM capacity. We present an example of LLaMA-65B in Section~\ref{sec:background-multi-turn}, which shows the KV caches fully occupy the free space within the HBMs in 14 seconds.
To address this challenge, \attname explores and exploits slower but larger-capacity storage hierarchies than HBMs, including host memory and disks, to provide adequate storage space for caching KV caches.

Thirdly, since disks have a much larger capacity than the host memory (tens of TBs v.s. several hundreds of GBs), most KV caches are retained in disks for \attname. As conversation requests arrive randomly, their corresponding KV caches are more likely to be located in disks, resulting in poor access performance.
To address this problem, \attname uses a scheduler-aware KV cache fetching scheme. This scheme pre-fetches the KV caches that are likely to be accessed from disks to the host memory, by utilizing the hints received from the inference job scheduler. When the free space of the host memory is not enough, \attname also adopts a scheduler-aware eviction scheme to efficiently identify the most suitable KV caches in memory and evict them to disks or out of the system.
 
Finally, when a conversation session surpasses the limit of the context window of LLMs, e.g., 4K in LLaMA-2~\cite{touvron2023llama2}, LLMs generally truncate the oldest tokens and limit the context to the most recent tokens~\cite{openai_truncation_tokens}. This truncation makes all saved KV caches of that conversation in \attname invalid since the positional information of all tokens embedded in the KV cache is changed. 
To overcome this issue, \attname decouples the positional encoding from the KV caches when saving them. It re-embeds the positional encoding into KV caches when loading them. After decoupling, truncation can be directly applied to the KV caches, thereby ensuring the reusability of the saved KV caches.

We implement the \attname and evaluate it using the real ShareGPT dataset~\cite{sharegpt_sharegptraw_2024}. Extensive experimental results demonstrate that \attname significantly decreases the time to the first token (TTFT) by up to 87\% and improves the prompt prefilling throughput by up to 7.8$\times$ for multi-turn conversations. It also reduces the end-to-end inference cost by up to 70\%. 
To summarize, this paper makes the following contributions: 
\begin{itemize}
\item We investigate the recomputation overheads of KV caches in LLMs across conversation turns and identify the challenges associated with retaining KV caches across multi-turn conversations.
\item We propose \attname, a new attention that allows the reuse of the KV caches for any ensuing conversation turns of the same session, achieving a significant reduction in the recomputation overhead of KV caches in LLMs.
\item To improve the efficiency of \attname, we design overlapped KV cache access, hierarchical KV cache placement, and positional encoding decoupled KV cache truncation schemes.
\item We thoroughly evaluate \attname with real datasets to demonstrate its efficacy and efficiency. 
\end{itemize}

%% file: tex/background.tex
\section{Background and Motivation} \label{sec:background}
This section first introduces the fundamentals of generative LLM inference and then explores the inefficiencies in LLMs during multi-turn conversations. 
The section finally discusses the design opportunities for dealing with these inefficiencies and the challenges faced in designing such a system.

\subsection{Generative LLM Inference Basics}
\textbf{Transformer Architecture.} 
The transformer has emerged as the widely accepted standard in generative LLM inference. The widely used LLMs like GPTs~\cite{Chatgpt} and LLaMAs~\cite{touvron2023llama, touvron2023llama2} are built upon the autoregressive transformer architecture~\cite{vaswani2017attention, han2021transformer}. During inference, these models process the prompt of the users and generate a response. The prompt is processed as a sequence of input tokens, and the response is generated by the model predicting the probability of subsequent tokens using the context of all the prior tokens. The transformer model consists of a chain of $l$ transformer layers, each consisting of two phases: \textit{self-attention} and \textit{feed-forward network} (FFN). 

For the input token list $X = [x_1, x_2,...x_s]$, each layer applies a series of projections on each token in $X$ using the weights $W_Q, W_K, W_V$. This generates the elements in the set of queries, keys, and values, referred to as $Q$, $K$, and $V$ respectively:
\begin{equation*}
    Q = W_QX, K = W_KX, V = W_VX
\end{equation*}
Subsequently, attention scores are computed via $Q$, $K$, and $V$:
\begin{equation*}
    Attention(Q, K, V) = softmax(\frac{QK^T}{\sqrt{d_K}})V 
\end{equation*}
where $\sqrt{d_K}$ is the dimension of the key vector $K$. Finally, the projection operation applies a linear transformation on attention scores. This projected result is handed to the FFN layer. The result from FFN is passed on to the next transformer layer as input. Finally, after the input has been processed through all $l$ transformer layers, the output is a probability vector that marks out the most probable output tokens.

\textbf{KV Cache:} The above process produces intermediate K and V tensors for each token. When generating subsequent tokens, all KV tensors of preceding tokens are necessary for computing the self-attention. These K and V tensors are generally cached in GPUs, referred to as the \textit{KV cache}. 
The KV cache typically has a large footprint. 
For example, GPT-3~\cite{Chatgpt, dettmers2022gpt3} generates a 4.5MB KV cache for each token.  
The size of KV cache linearly increases with the number of prompt tokens. A conversation session containing thousands of tokens will produce several GBs of KV cache.

\subsection{Autoregressive Generation}

\begin{figure}[t]
        
     \begin{subfigure}[h]{0.2\textwidth}
         \includegraphics[width=1.5in]{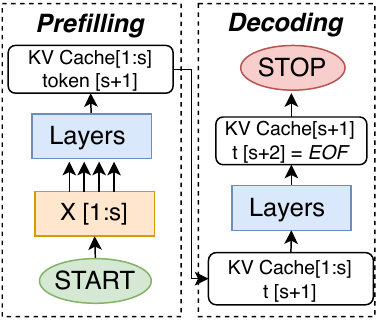}
        \caption{Two-phase illustration.} 
        \label{fig:two_phase}
     \end{subfigure}
    \hfill
     \begin{subfigure}[h]{0.25\textwidth}
         \includegraphics[width=1.65in]{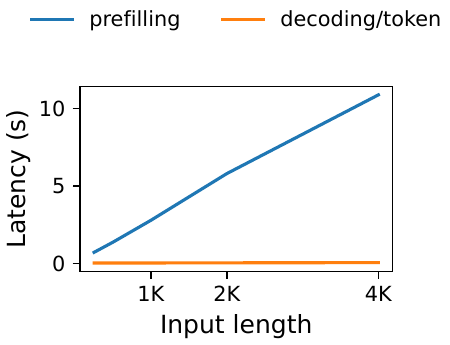}
         \caption{Execution latency.}
         \label{fig:prefill_decoding}
     \end{subfigure}
     \caption{Prefilling and decoding phases. Latency measured for LLaMA-70B of batch size 8 on 4 A100 GPUs.}
     \vspace{-10pt}
\end{figure}

As illustrated in Figure~\ref{fig:two_phase}, transformer-based generation can logically be identified as two distinctive phases\cite{agrawal2023sarathi}. 

\textbf{The prefilling phase.} 
Given a request prompt, the generation takes the prompt token list $X = [x_1, x_2, ... x_s]$ as input and then proceeds to compute the token $x_{s+1}$. This process generates a series of KVs, specifically forming the KV cache ranging from 1 to $s$, which are used for the decoding phase.     

\textbf{The decoding phase.}
The decoding phase iteratively generates output tokens. 
The decoding phase takes token $\textit{s+}1$ and the KV cache $[1$:$\textit{s}]$ from the prefilling phase as input to compute the KV cache $\textit{s+}1$ and the token $\textit{s+}2$. 
The generation process iteratively continues until the generated token is <\textit{eos}> or the iteration number reaches the maximum allowed generation number. The decoding phase happens sequentially due to the heavy data dependency on the previous iteration. 

The two phases present distinct execution time characteristics. The prefilling phase, computing the KV cache in parallel, has a duration closely tied to the number of input prompt tokens. As shown in Figure~\ref{fig:prefill_decoding}, the execution time of the prefilling phase increases with more input tokens. In contrast, the decoding phase computes for a single token per iteration, resulting in relatively constant computation time per iteration. 

\begin{figure}[t]
    \centering
    \includegraphics[width=3.2in]{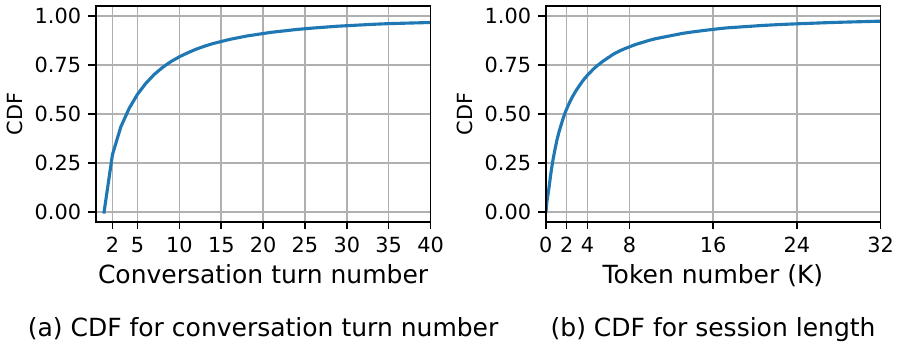}
    \caption{(a) Distribution for conversation turn number in ShareGPT~\cite{sharegpt_sharegptraw_2024}. (b) The session length distribution of ShareGPT. For better display effect, the statistics exclude conversations with over 40 turns or sessions that exceed a length of 32K.}
    \begin{subfigure}{\textwidth}
    \centering
    \phantomcaption\label{fig:dist_turnnum}
    \phantomcaption\label{fig:dist_tokennum}
  \end{subfigure}
    \label{fig:multiturn-analysis}
\end{figure}

\subsection{Multi-turn Conversation Inference}
\label{sec:background-multi-turn}

\begin{figure}[t]
    \centering
    \includegraphics[width=3.4in]{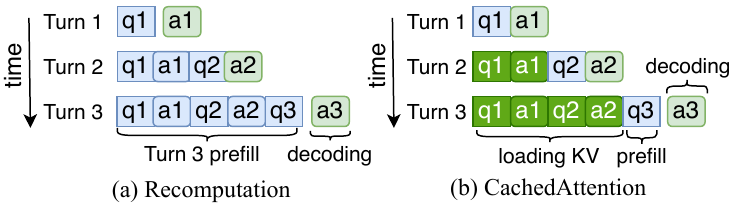}
    \caption{Comparison of recomputation and \attname.}
    \vspace{-20pt}
    \begin{subfigure}{\textwidth}
    \centering
    \phantomcaption\label{fig:conversation}
    \phantomcaption\label{fig:cachedattention}
  \end{subfigure}
    \label{fig:baseline_vs_cachedattention}
\end{figure}

Engaging humans in multi-turn conversations is a fundamental feature of modern LLMs. 
A multi-turn conversation session consists of a series of continuous conversations, denoted as $D = [d_1, d_2, ... d_N]$. In each conversation $d_j$, a user inputs a new question or command $q_j$ and then awaits the response $a_j$ from the LLM. 
To maintain a coherent context and understanding of the conversation session, the LLM generates $a_{N+1}$ based on both the historical tokens from all previous conversation turns $d[1$:$N]$ and the input tokens of the current turn, denoted as $q_1a_1q_2a_2...q_Na_Nq_{N+1}$. 

\begin{figure}[t]
    \centering
    \includegraphics[width=3.45in]{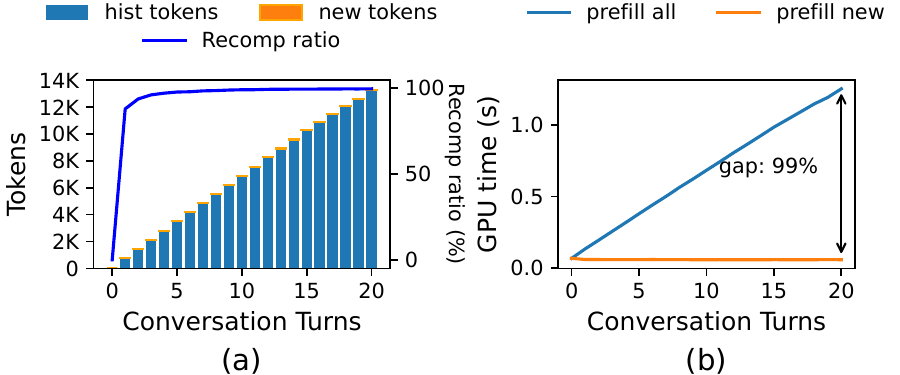}
    \vspace{-20pt}
    \caption{Recomputation inefficiencies. (a) Average numbers of historical tokens and new tokens in different turns of ShareGPT~\cite{sharegpt_sharegptraw_2024}. (b) GPU time for prefilling all tokens and only new input tokens in ShareGPT with Mistral-7B~\cite{jiang2023mistral} on 1 A100 GPU.}
    \vspace{-20pt}
    \begin{subfigure}{\textwidth}
    \centering
    \phantomcaption\label{fig:recompute-tokens}
    \phantomcaption\label{fig:recompute-costs}
  \end{subfigure}
  
    \label{fig:enter-label}
\end{figure}

Based on the analysis of ShareGPT~\cite{sharegpt_sharegptraw_2024, sharegpt_sharegpt_2024}, a real dataset collected from ChatGPT that includes more than 90K conversations, we observe that 73\% of conversations are multi-turn, as shown in Figure~\ref{fig:dist_turnnum}. Moreover, 30\% of conversations have more than 4K tokens as shown in Figure~\ref{fig:dist_tokennum}.

However, executing multi-turn conversations in current LLM serving engines is inefficient due to the repetitive computation of KV caches across multiple conversation turns. 
As shown in Figure~\ref{fig:conversation}, in conversation turn 1, the LLM serving engine generates the KV cache of $q_1$ and $a_1$. After finishing turn 1, the LLM serving engine discards the KV cache to reclaim the HBM space. In turn 2, the LLM serving engine re-generates the KV cache of $q_1$ and $a_1$. In turn~3, the KV cache of $q_1$, $a_1$, $q_2$, and $a_2$ is re-generated.
As the session expands, the historical tokens keep accumulating and the amount of repetitive computation significantly increases. 
As shown in Figure~\ref{fig:recompute-tokens}, with the increase of the conversation turns, the percentage of historical tokens will be more than 99\% in a new conversation. The repetitive computation time occupies 99\% of the prefilling time (a.k.a., time to the first token) in the new conversation, as shown in Figure~\ref{fig:recompute-costs}.

\subsection{Opportunities and Challenges} \label{sec:challenge}
Based on the analysis above, we observe that if the KV caches can be reused across multiple turns of conversations, up to 99\% of prefilling cost can be reduced. 
Specifically, the KV caches of historical conversations can be saved in a KV caching system out of GPUs. 
Upon the reactivation of a conversation session, GPUs load the associated KV caches from the KV caching system and reuse them for the new-turn conversation. 
Nevertheless, to build an efficient KV caching system, there exist many significant challenges.

\textit{\textbf{1) High KV cache access overheads.}} 
During the inference, the computation of GPUs can be blocked due to waiting for the KV caches to be loaded from the KV caching system. The block time is non-negligible compared to the repetitive computation time of the KV cache, making the KV caching solution lose efficacy.
For example, we evaluate the inference time of the LLaMA-65B model using 4 NVIDIA A100 GPUs and observe that prefilling 2K tokens of a prompt consumes about 360~ms. In contrast, loading the KV cache of the 2K tokens (5~GB) from host memory to GPUs consumes about 192~ms (the GPU system with 16 lanes of PCIe Gen4 has about 26~GB/s of effective data transmission bandwidth).

\textit{\textbf{2) High storage capacity requirement of KV caches.}} 
Storing KV cache for each request consumes a substantial amount of storage space. 
For instance, when using 4 A100 GPUs each with 80GB HBM to run LLaMA-65B, prefilling 2K tokens consumes about 360~ms. This process generates 5~GB of KV cache, indicating the generation speed of the KV cache is about 13.9~GB/s. As 130~GB of HBM space is allocated to store the model, the remaining 190~GB of free HBM space will be fully occupied by the KV cache within 14 seconds. If spilling the KV cache to the host memory (e.g., 512~GB space), the host memory will be filled in less than 1 minute. Using disks to save the KV cache can extend the storage space but incur worse access performance, as presented below.

\textit{\textbf{3) Suitable placement of KV caches in different hierarchies.}} 
Disks provide much larger capacity than the host memory (tens of TBs v.s. several hundreds of GBs). Thus most KV caches are retained in disks. However, the disks have an access bandwidth of less than 5~GB/s. As conversation requests arrive randomly, their corresponding KV caches are more likely to be located in disks when being accessed, resulting in poor inference performance. It is essential to ensure that the KV cache to be accessed in the immediate future is always placed in the host memory instead of disks.

\textit{\textbf{4) Unexpected invalidation of the saved KV caches.}}
With the number of conversation turns increasing, the historical tokens can exceed the context window limitation. LLM serving engines generally perform token truncation~\cite{openai_truncation_tokens, han2023lminfinite} to reduce the input prompt. The truncation has no impact on previous LLM serving engines since they always recompute the KV cache from the truncated input prompt. However, the truncation makes the KV caches saved in \storagename invalid, since the position of each token is changed after truncation. Thus it cannot match the old embedded positional encoding in the saved KV cache. Such context window overflow can occur with a high probability.
As shown in Figure~\ref{fig:dist_tokennum}, 47\% and 30\% of conversation sessions have a context longer than 2K and 4K, respectively. It means that when using the LLaMA-2 family with 4K context window \cite{touvron2023llama2}, the context window overflow occurs in 30\% of conversation sessions. When using the OPT family with 2K context window \cite{zhang2022opt}, the context window overflow occurs in 47\% of conversation sessions.

%% file: tex/design.tex
\section{The \attname Design}
\label{sec:design}

\subsection{Overview}
In this paper, we propose a new attention mechanism, called \textit{\attname}, which enables the reuse of KV caches across multi-turn conversations. Unlike the conventional attention mechanism that uses all prompt tokens for prefilling, \attname uses the new tokens input in the new conversation turn and the KV caches of historical tokens for prefilling, as shown in Figure~\ref{fig:baseline_vs_cachedattention}. Specifically, when the associated conversation session is inactive, \attname saves the KV cache in a KV caching system, named \textit{\storagename}, instead of discarding them as in the conventional attention mechanism. If the same conversation session is activated in the future, its KV cache is fetched from \storagename and reused for inference. 
By doing so, \attname only executes the prefilling of partial prompt tokens, i.e., the new tokens input in the new turn of conversation, rather than prefilling all prompt tokens. 
As shown in Figure~\ref{fig:cachedattention}, when executing the inference of Turn 3, the KV cache of $q_1$, $a_1$, $q_2$, and $a_2$ is reused and only $q_3$ needs to be prefilled. \attname effectively eliminates the repetitive computation overhead of the historical tokens, thereby reducing the prefilling cost.

\begin{figure}[t]
    \centering
    \includegraphics[width=3.2in]{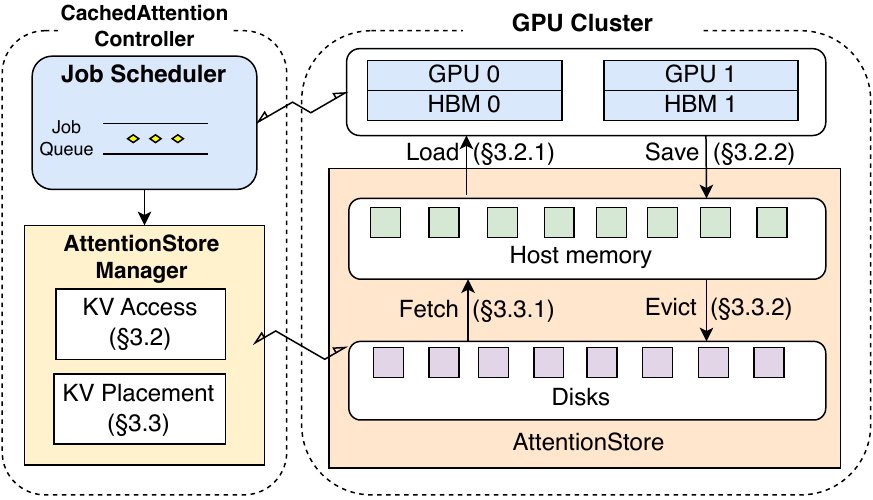}
    \caption{The system architecture of \attname.}
    \vspace{-15pt}
    \label{fig:deepdia-arch}
\end{figure}

Figure~\ref{fig:deepdia-arch} shows the architectural overview of \attname. It maintains a hierarchical KV caching system, i.e., \storagename, with efficient KV cache access, placement, and truncation techniques to address the challenges mentioned in Section~\ref{sec:challenge}.

For Challenge 1, to reduce the overhead of KV cache loading from \storagename into HBMs, \attname leverages a layer-wise pre-loading scheme to overlap the KV cache loading with the inference computation. To reduce the KV cache saving overhead from HBMs to host memory, \attname leverages an asynchronous saving scheme to overlap the saving with the inference computation. (\S \ref{sec:kv-transmission}). 

For Challenges 2 and 3, to enlarge the available storage space for caching KV caches, \attname employs multi-tier cost-effective storage mediums in \storagename, i.e., host memory and disks. To reduce the impact of accessing slow disks on the inference performance, we present a scheduler-aware fetching scheme that leverages the hints from the job scheduler to prefetch KV caches to be accessed from disks to host memory. Meanwhile, to efficiently leverage the limited host memory space, we present a scheduler-aware eviction scheme that identifies the least valuable KV caches and evicts them to disks or out of the caching system (\S \ref{sec:kv-caching}). 

For Challenge 4, to deal with the invalidation of KV caches saved in \attname due to context window overflow, we utilize a positional encoding decoupled truncation scheme to save the KV caches without positional encoding embedded, and hence support the truncation directly on KV caches. When loading the KV cache, \attname re-embeds the new positional encoding into the KV caches (\S \ref{sec:kv-invalidation}).

\subsection{Overlapped KV Cache Access} \label{sec:kv-transmission}
The use of slower memory/storage hierarchies results in significant access overhead because KV caches need to be transferred between HBMs and the slower mediums, blocking the inference and causing a waste of computational resources. To reduce the KV cache loading overheads from host memory to HBMs, \attname uses a layer-wise pre-loading scheme to overlap the loading of the KV cache with the inference computation layer by layer (\S \ref{sec:kv-retrieval}). To reduce the KV cache saving overheads, \attname develops an asynchronous saving scheme that overlaps the saving of KV caches with the inference computation (\S \ref{sec:kv-retention}).

\subsubsection{Layer-wise Pre-loading from Memory to HBMs} \label{sec:kv-retrieval}

\begin{figure}[t]
\centering
\begin{subfigure}{1\linewidth}
\centering
\includegraphics[width=1\linewidth]{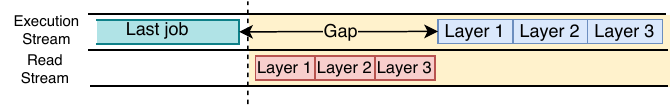}
\caption{Baseline: KV cache loading without concurrent operations.}
\label{fig:kv-retrieval-baseline}
\end{subfigure}%
\\
\begin{subfigure}{1\linewidth}
\centering
\includegraphics[width=1\linewidth]{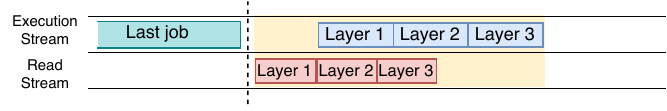}
\caption{Layer-wise pre-loading without buffer.}
\label{fig:kv-retrieval-prefetching}
\end{subfigure}
\\
\begin{subfigure}{1\linewidth}
\centering
\includegraphics[width=1\linewidth]{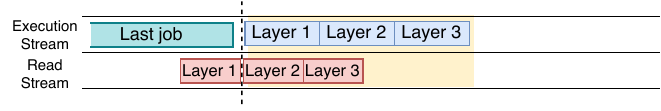}
\caption{Layer-wise pre-loading with buffer.}
\label{fig:kv-retrieval-buffer}
\end{subfigure}

\caption{Layer-wise KV cache pre-loading. Blue blocks indicate the execution of each transformer layer. Red blocks indicate the KV cache loading of each transformer layer.}
\vspace{-10pt}
\label{fig:figure}
\end{figure}

\begin{figure}[t]
\centering
\begin{subfigure}{1\linewidth}
\centering
\includegraphics[width=1\linewidth]{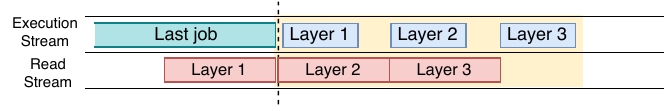}
\caption{Layer-wise pre-loading with imperfect overlapping.}
\label{fig:kv-retrieval-imperfect}
\end{subfigure}
\\
\begin{subfigure}{1\linewidth}
\centering
\includegraphics[width=1\linewidth]{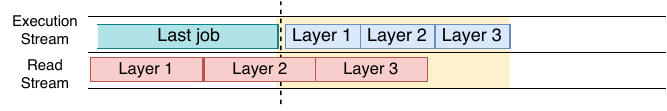}
\caption{Perfect pre-loading with a customized larger buffer.}

\label{fig:kv-retrieval-large-buffer}
\end{subfigure}
\caption{Layer-wise KV cache pre-loading.}
\vspace{-10pt}
\label{fig:figure}
\end{figure}

\attname loads KV caches from the host memory to HBMs, resulting in high data access overhead. The access process is in the critical path of the inference execution as shown in Figure~\ref{fig:kv-retrieval-baseline}, since GPUs must rely on the KV cache to execute the inference computation.  This overhead becomes more significant as the size of the KV cache increases, as discussed in Section~\ref{sec:challenge}. To eliminate this overhead, \attname employs a layer-wise pre-loading scheme to mitigate the impact. The main idea is to overlap the loading of the KV cache with the prefilling computation of new input tokens for the conversation. In particular, the LLM model is chained by multiple transformer layers, each with its own KV cache. As the GPU executes a layer, the KV cache needed by the subsequent layers can be loaded from the host memory concurrently. By doing so, when the GPU starts computing the self-attention for a layer, the corresponding KV cache of the layer is already in the HBM execution buffer.

Figure~\ref{fig:kv-retrieval-prefetching} shows how the layer-wise pre-loading scheme overlaps the KV cache fetching time with the computation time, using a 3-layer model for simplicity. 
Before initiating the computation of Layer 1, the KV cache for this layer must first be prepared in the HBM. The read stream first issues a KV cache loading operation to read the KV cache for Layer 1 into the HBM execution buffer. The execution stream then starts computing Layer 1. While the execution stream is computing one layer, the read stream concurrently loads the KV cache for the next layers, thereby overlapping loading with computation. However, a gap still exists between the last job and the first layer of the current job, since the loading can only commence once the HBM execution buffer is available, i.e., the last job is finished. To further mitigate the gap between the last job and the first layer of the current job, \attname reserves an HBM read buffer to eliminate the gap. Specifically, as shown in Figure~\ref{fig:kv-retrieval-buffer}, with the read buffer, the read stream doesn't have to wait for the release of the execution buffer from the last job. The read stream can start the pre-loading while the last job is running.

However, pre-loading may fail to fully overlap with the computation if the KV cache loading time is longer than the prefilling computation time. As shown in Figure~\ref{fig:kv-retrieval-imperfect}, multiple gaps exist between the computation of layers because the KV cache fetching time for each layer exceeds the computation time for each layer, resulting in imperfect overlapping. The overhead can be further minimized by employing a customized larger pre-loading buffer. With the larger buffer, pre-loading can be issued with an earlier start. For instance, as shown in Figure~\ref{fig:kv-retrieval-large-buffer}, with the larger buffer, pre-loading is allowed to pre-load KV cache for more layers and thus the gaps between layers can be overlapped. 
Let $T_{load}$, $T_{pref}$, $L_{hist}$ and $L_{new}$ denote the access time of the KV cache for a token, the prefilling time for a token, the length of historical tokens in a session, and the length of new input tokens in the conversation, respectively. Imperfect overlapping happens when $T_{load}L_{hist} > T_{pref}L_{new}$, which indicates that the transmission time is larger than the partial prefilling time. The buffer is used to fill up the time gap $T_{load}L_{hist} - T_{pref}L_{new}$. Combined with the PCIe bandwidth $B$, the buffer size can be set by the following formula:
    $S_{buf} = B(T_{load}L_{hist} - T_{pref}L_{new}).$

\subsubsection{Asynchronous Saving from HBMs to Memory} \label{sec:kv-retention}

\begin{figure}[t]
\centering
\begin{subfigure}{1\linewidth}
\centering
\includegraphics[width=1\linewidth]{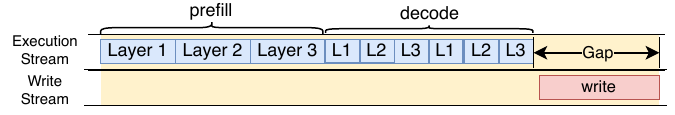}
\caption{Baseline: KV cache saving without concurrent operations.}

\label{fig:kv-retention-baseline}
\end{subfigure}%
\\
\begin{subfigure}{1\linewidth}
\centering
\includegraphics[width=1\linewidth]{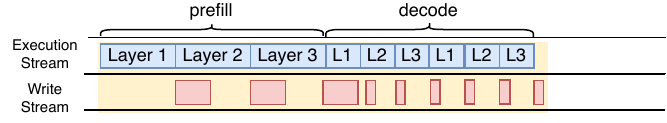}
\caption{Asynchronous KV cache saving with overlapping.}
\label{fig:kv-retention-overlap}
\end{subfigure}
\caption{Asynchronous KV cache saving.}
\vspace{-10pt}
\label{fig:figure}
\end{figure}

\attname needs to save KV caches to host memory to enable the reuse of the KV caches across conversations. A baseline method to save the KV caches is to write all produced KV caches together after the round of conversation ends. This method however potentially delays the execution of the next scheduled jobs since the KV saving time is on the critical path of inference, as shown in Figure~\ref{fig:kv-retention-baseline}. To reduce this overhead, \attname incorporates an asynchronous KV cache saving scheme to overlap the KV cache write-back with the computation, which also considers the different characteristics of prefilling and decoding phases to perform different overlapping mechanisms. 

Specifically, 
the generation speeds of KV caches at the prefilling and decoding phases are different.
The prefilling phase processes tokens concurrently, thus generating substantial volumes of KV cache within a restricted timeframe. In contrast, the decoding phase generates the KV cache of one token at a time. 
As shown in Figure~\ref{fig:kv-retention-overlap}, for the prefilling phase, as each self-attention operation can produce a significant amount of KV cache, the write stream retains the KV cache layer by layer. The KV cache produced by the prefilling phase can be overlapped with the decoding phase. For the decoding phase, as the KV cache is iteratively produced, the write stream writes back the KV cache layer by layer while decoding. To avoid getting stuck if the KV cached is not fully written back when the decoding is already finished, we also reserve an HBM write buffer to cover such cases similar to the read buffer used in the KV cache prefetching. The unfinished KV caches are temporarily moved to the write buffer to avoid blocking the execution of the next job.

\subsection{Hierarchical KV Cache Placement} \label{sec:kv-caching}
\attname leverages both host memory and disks to expand the available space for KV cache storage. The access speed of host memory, i.e., DRAM, is much higher than disks, i.e., SSDs, (tens of GB/s v.s. several GB/s). If the KV caches to be accessed are always found in the host memory instead of disks, the access performance of KV caches will be optimal. To achieve this, \attname applies a scheduler-aware fetching scheme to pre-fetch the KV caches from disks to host memory, ensuring KV cache access at the optimal speed (\S \ref{sec:kv-migration}), and a scheduler-aware eviction scheme to evict suitable KV caches from host memory to disks (\S \ref{sec:kv-eviction}).

\subsubsection{Scheduler-aware Fetching from Disks to Memory} \label{sec:kv-migration}

Since disks have a much larger capacity than the host memory (tens of TBs v.s. several hundreds of GBs), most KV caches are retained in disks for \attname. As conversation requests arrive randomly, their corresponding KV caches are more likely to be located in disks, resulting in poor access performance.

To address the problem, we leverage a scheduler-aware KV cache fetching scheme to pre-fetch the KV caches to be accessed from disks to the host memory. This is done by utilizing the hints from the inference job scheduler. Specifically, the job scheduler maintains a job queue, thus having the full knowledge of waiting jobs. \attname applies a look-ahead prefetching window to watch for the waiting jobs to be executed. If the KV cache of the waiting jobs is hit in the disks, \attname will pre-fetch the KV cache of waiting jobs from the disks to host memory before these waiting jobs are executed. The length of the look-ahead prefetching window is determined by the available capacity in the host memory. Given the available memory capacity for prefetching $C_{mem}$ and the average KV size of a session $S_{kv}$, the prefetching window length is $L_{pw} = C_{mem} / S_{kv}$.

\begin{figure}[t]
    \centering
    \includegraphics[width=3.2in]{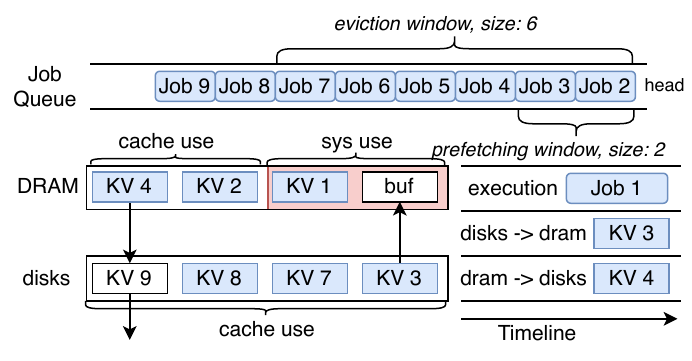}
    \caption{Scheduler-aware KV cache fetching and eviction.}
    \vspace{-10pt}
    \label{fig:migration-algo}
\end{figure}

A scheduler-aware fetching example is shown in Figure~\ref{fig:migration-algo}. As Job 1 is executing, the KV cache manager applies a look-ahead window size of 2 (the host memory has 2 KV cache slots for the KV cache fetching) to check the KV cache hit status of the waiting Jobs 2-3. The KV cache for Job 2 is hit in the host memory but the KV cache for Job 3 is not in the host memory. Then the KV cache fetching threads start fetching the KV cache for Job 3 from disks to the host memory.  

Note that \attname includes a host memory buffer that allows for seamless fetching of KV caches from disks to memory, preventing any delays when the host memory is full. When the capacity of the free memory reaches a defined threshold, \attname triggers a KV eviction from host memory to disks to ensure the constant availability of the host memory buffer. The eviction process from host memory to disks is presented in the next subsection.

\subsubsection{Scheduler-aware Eviction from Memory to Disks} \label{sec:kv-eviction}
When the free space in the host memory is exhausted, we need to evict some KV caches from the host memory to disks. Meanwhile, if the disks are full, we also need to evict some KV caches stored in the disks out of the system. Therefore, it is important to carefully choose the suitable KV cache candidates to be evicted for achieving a high cache hit rate.

Different from the existing cache eviction strategies, such as the least-recently-used (LRU)~\cite{touzeau2019fast}, first-in-first-out (FIFO)~\cite{dan1990approximate}, and their variants, which solely rely on the historical access information of the KV caches, \attname presents a scheduler-aware eviction scheme which can leverage the future access information of KV caches to achieve a higher cache hit rate. The job queue in the job scheduler gives us the opportunity to achieve this. Specifically, \attname maintains a look-ahead eviction window in the job queue. The maximum length of the look-ahead eviction window is determined by the total storage capacity of \storagename. Assume the total available capacity in the disks is $C_{disk}$. The look-ahead eviction window length is $(C_{mem} + C_{disk}) / S_{kv}$.   When \attname attempts to evict one item out of \storagename, if finding the item to be evicted in the look-ahead eviction window, the item is exempted. When \attname evicts one item from the host memory to disks, the item located at the tail of the look-ahead eviction window has a higher priority to be evicted. Note that one item corresponds to all KV caches associated with a conversation session, which is the minimal eviction and fetching granularity in \attname. This is because the KV cache in the same conversation session is either all used or none of it is used.

A scheduler-aware eviction example is shown in Figure~\ref{fig:migration-algo}. When the KV cache of Job 3 is chosen to be migrated to the host memory, the buffer will be utilized. To maintain a buffer in the host memory, \attname needs to evict KV caches from the host memory to the disks. \attname employs a look-ahead eviction window of size 6 to monitor the KV cache status of the jobs. First, it finds that the KV caches in the host memory all have an associated job in the job queue. It then proceeds to scan the look-ahead eviction window from tail to head, prioritizing the eviction of jobs near the tail. Therefore, the KV cache for Job 4 is selected to be evicted from the host memory to the disks. Since the disks are also full, the scanning process identifies that the last arrived Job 9 in the job queue is the most suitable candidate to be evicted. Finally, the KV cache for Job 4 is moved to the location previously occupied by Job 9.

\subsection{Decoupled KV Cache Truncation} 
\label{sec:kv-invalidation}

\begin{figure}
    \centering
    \includegraphics[width=3.2in]{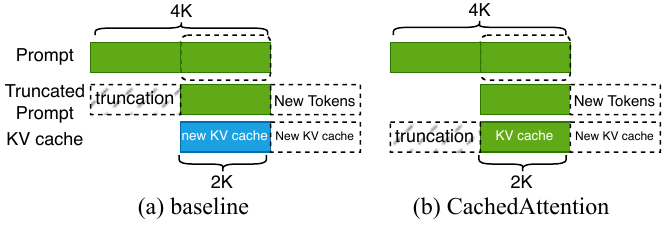}
    \vspace{-10pt}
    \caption{Illustration of managing context window overflow. Context window size: 4K, truncation ratio: 2K. 
    (a) Baseline. Token truncation\cite{openai_truncation_tokens}.
    (b) KV cache truncation. 
    }
    \begin{subfigure}{\textwidth}
    \centering
    \phantomcaption\label{fig:token-truncation}
    \phantomcaption\label{fig:kv-cache-truncation}
  \end{subfigure}
  \vspace{-20pt}
\end{figure}
When the historical tokens exceed the limitation of the context window, LLM serving engines generally perform token truncation\cite{openai_truncation_tokens}. As shown in Figure~\ref{fig:token-truncation}, the context window size is 4K. Once the context window overflows, the LLM serving engines cut off the first 2K tokens of the prompt. The truncation has no impact on previous LLM serving engines since they always recompute the KV cache based on the input prompt, regardless of truncation. However, the truncation makes the KV caches stored in \attname invalid, significantly reducing the efficiency of \attname.  This is due to the positional encoding embedded in the KV caches. On performing token truncation on the prompt, the position of each token is changed. The positional encoding embedded in the KV caches cannot be modified to match the positions of tokens in the prompt, making the KV caches invalid.

\begin{figure}
    \centering
    \includegraphics[width=3.4in]{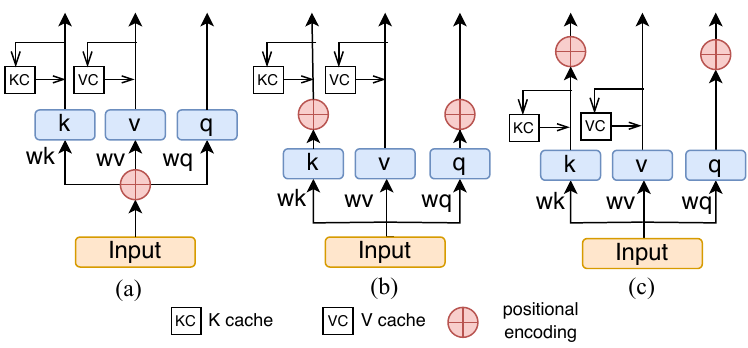}
    \vspace{-20pt}
    \caption{(a) Absolute positional encoding. (b) Relative positional encoding. (c) KV cache with decoupled positional encoding.
    }
    \vspace{-15pt}
    \begin{subfigure}{\textwidth}
    \centering
    \phantomcaption\label{fig:absolute}
    \phantomcaption\label{fig:relative}
    \phantomcaption\label{fig:decouple}
  \end{subfigure}
\end{figure}

To address this problem, \attname enables the KV caches after truncation to be still valid via decoupling the positional encoding. \attname needs to work with the relative position encoding (RPE)~\cite{su2024roformer, xiao2023efficient, touvron2023llama}. Unlike the absolute positional encoding (APE) in which positional encodings are added to the input, RPE directly embeds the positional encodings in the query (Q) and key (K) vectors, as shown in Figure~\ref{fig:relative}. Extensive research shows that RPE allows LLMs to learn from longer data sequences than APE~\cite{vaswani2017attention, devlin2018bert, zhang2019ernie}. Therefore, RPE is widely used in modern LLMs, e.g., LLaMA~\cite{touvron2023llama}, T5~\cite{xue2020mt5}, Falcon~\cite{refinedweb}, Mistral~\cite{jiang2023mistral}, Mixtral~\cite{jiang2024mixtral} and Transformer-XL~\cite{dai2019transformer}.
By simply moving the time of caching KVs before embedding positional encodings in RPE as shown in Figure~\ref{fig:decouple}, \attname can store the KVs without embedded positional encodings in \storagename. When reusing the KVs in \attname, the KVs are embedded with the new positional encodings and further used for the following inference.

\begin{figure}
    \centering
    \includegraphics[width=3.0in]{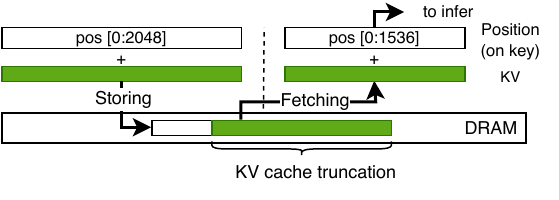}
    \vspace{-10pt}
    \caption{Illustration of KV cache truncation with \attname. 
    }
    \vspace{-15pt}
    \label{fig:cachedattention-truncation}
\end{figure}

Figure~\ref{fig:cachedattention-truncation} provides an example of how \attname supports KV cache truncation. \attname stores the KV cache without the positional encodings. In the cases where KV cache truncation becomes necessary, the LLM engine retrieves the truncated KV cache (i.e., KV [0:1536]) and loads it to the HBM. The new positional encodings are subsequently applied to the KV cache. 

Note that \attname also allows for selective preservation of certain KV cache for compression, e.g., the initial tokens with important scores~\cite{xiao2023efficient} or important tokens \cite{ge2023model, zhang2023h, liu2023scissorhands}, to further improve the generation quality of LLMs. Specifically, a given KV cache compression technique essentially provides a methodology for creating a token discarding list (TDL), in the prompt, e.g., discarding the least important tokens. \attname straightforwardly complies with the TDL, discarding the KV cache associated with the TDL within the \storagename. After discarding the selected KV caches, \storagename delivers the pruned KV cache to the inference engine upon an inference execution.

%% file: tex/evaluation.tex
\begin{figure*}[h]
\centering
\setkeys{Gin}{width=1\linewidth}
\begin{minipage}[t]{0.24\linewidth}
\includegraphics[width=\linewidth]{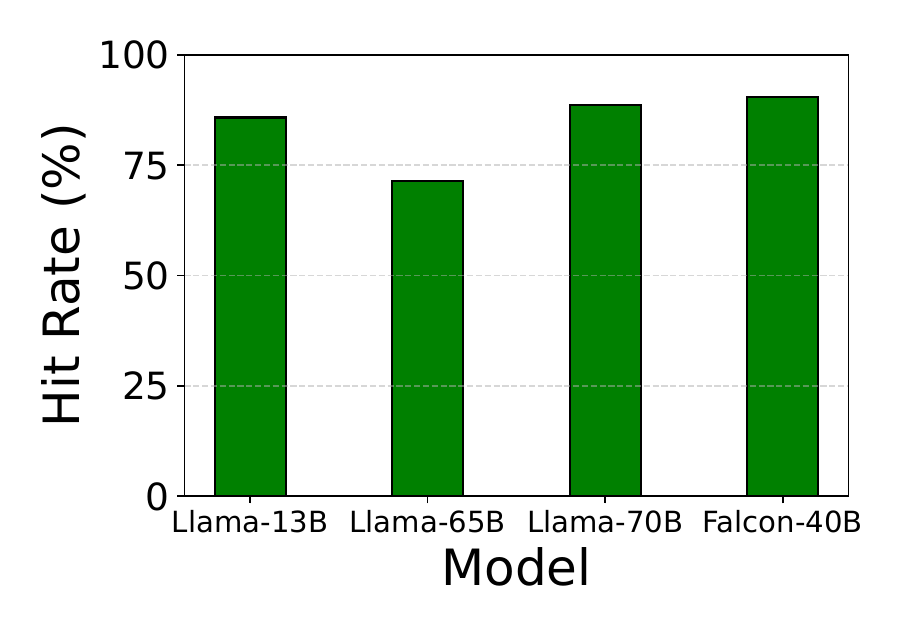}
\vspace{-25pt}
\caption{Cache hit rate.} \label{fig:e2e-hit-ratio}
\end{minipage}\hfill
\begin{minipage}[t]{0.24\linewidth}
\includegraphics[width=\linewidth]{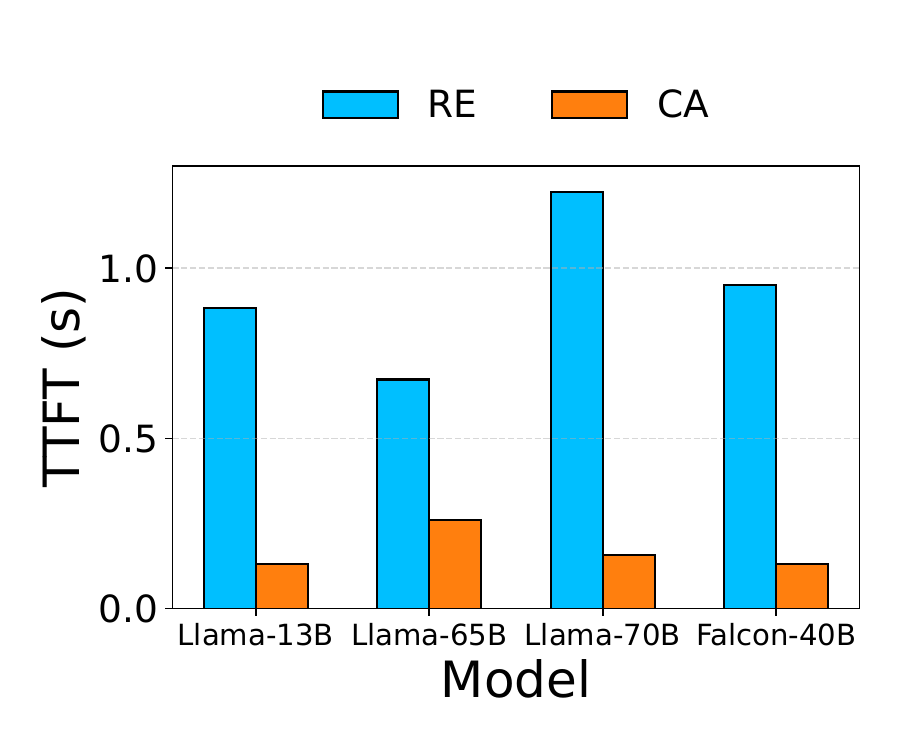}
\vspace{-25pt}
\caption{Time to first token.} \label{fig:e2e-ttft}
\end{minipage}\hfill
\begin{minipage}[t]{0.24\linewidth}
\includegraphics[width=\linewidth]{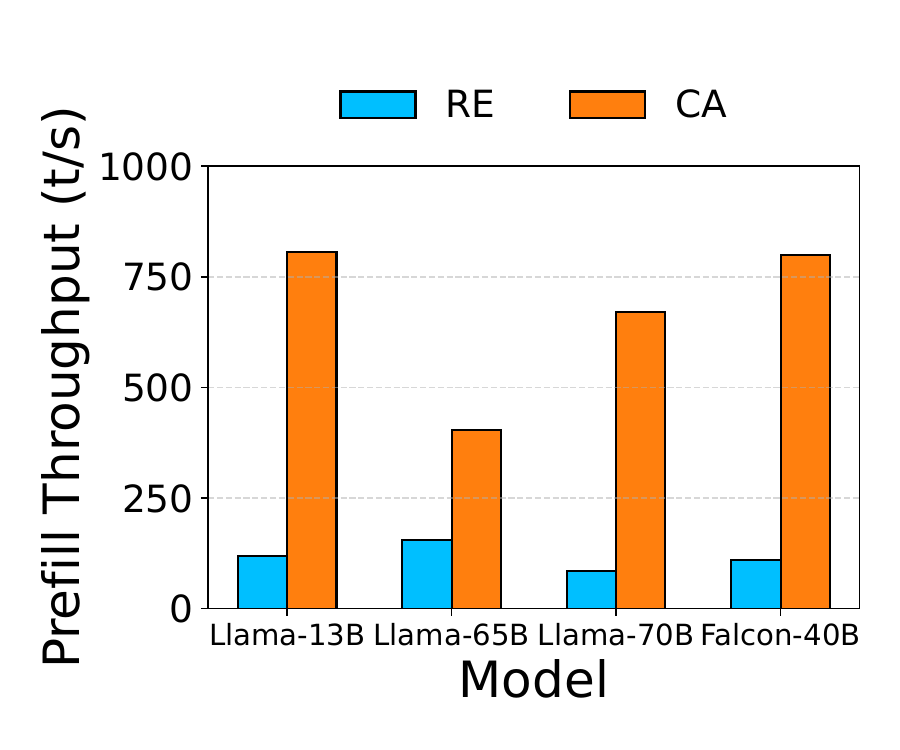}
\vspace{-25pt}
\caption{Prefill throughput.} \label{fig:prefill_throughput}
\end{minipage}\hfill
\begin{minipage}[t]{0.24\linewidth}
      \includegraphics[width=\linewidth]{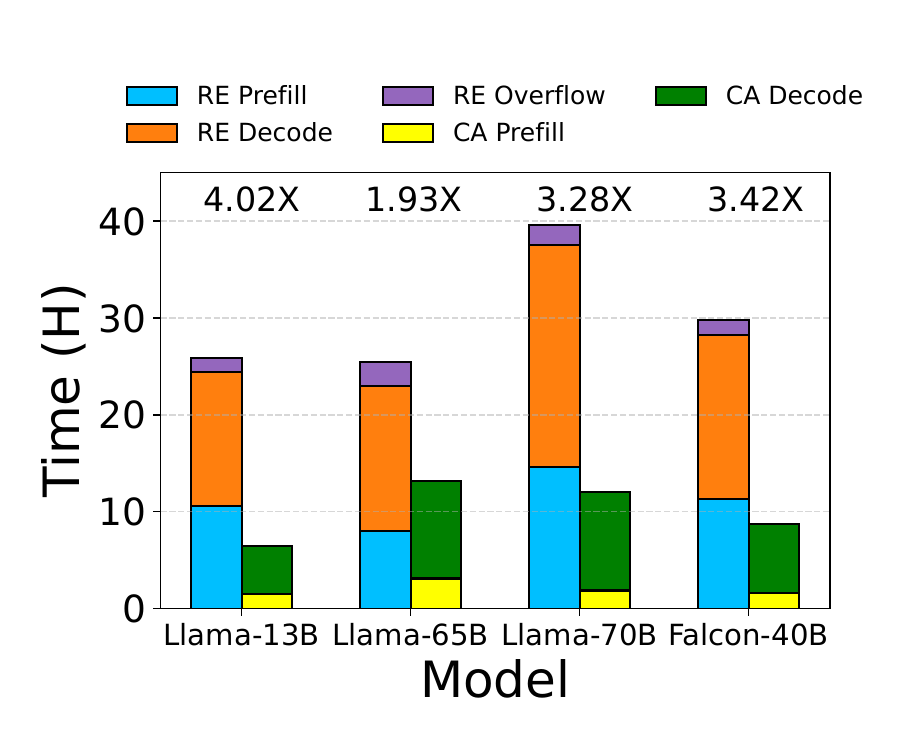} 
      \vspace{-25pt}
    \caption{GPU time.} \label{fig:e2e-performance}
\end{minipage}
\vspace{-10pt}
\end{figure*}
\section{Performance Evaluation}
\label{sec:evaluation}

\subsection{Experimental Setup}
\textbf{Testbeds.} All our experiments are performed on 4 NVIDIA A100 GPUs, each with 80GB HBM. The system is equipped with 128GB DRAM and 10TB SSDs. GPUs are connected to the host via PCIe Gen 4. 

We implement \attname in Pytorch and Python. 
The host memory and disks are managed in the form of blocks to improve storage utilization, similar to \cite{kwon_efficient_2023}. Our internal storage allocator allocates and deallocates storage blocks on demand. For the model executor, \attname integrates the implementation of popular LLMs such as LLaMA~\cite{touvron2023llama} and Falcon~\cite{refinedweb} based on Transformers~\cite{wolf2019huggingface}. Dedicated CUDA streams are used for moving data between the GPUs and the host memory, overlapping the computation with proactive swapping. Separate IO threads migrate data between the host memory and the disks, overlapping the execution with the KV cache migrations. Continuous batching~\cite{yu_orca_2022} is enabled through experiments.

\textbf{Models.} The experiments evaluate the open-sourced  LLaMA-1 model with 65B~\cite{touvron2023llama}, LLaMA-2 models~\cite{touvron2023llama2} with 13B, 70B, and Falcon 40B~\cite{refinedweb}. The intermediate activation uses FP16, aligned with prior systems~\cite{yu_orca_2022, llm_vllm_2023}. We also implement Mistral-7B~\cite{jiang2023mistral} with a 32K context window. Unless specified otherwise, LLaMA-13B operates on two GPUs with 24 batches, while LLaMA-65B, LLaMA-70B, and Falcon-40B run on four GPUs, handling 24 batches each. 

\textbf{Workloads.} The workload is integrated from the ShareGPT dataset~\cite{sharegpt_sharegptraw_2024, zheng2023judging}. As there is no public request arrival timestamp available in the dataset, we generate request arrival times based on the Poisson distribution with various arrival rates, following prior works~\cite{kwon_efficient_2023, wu_fast_2023}. We set the number of different sessions arriving per second according to a Poisson distribution (with $\lambda$ = 1.0). 9K conversation sessions are used in the experiments.

\textbf{Baseline.} We compare \attname (CA) with re-computation (RE). RE only keeps historical tokens of conversation sessions. It discards KV caches after serving a conversation and does not keep the KV cache while the conversation session is inactive. When a conversation associated with a particular session becomes active again, RE leverages the historical tokens from that session to recompute their KV caches. When the historical tokens exceed the context window limitation, RE applies token truncation, same as the general LLM services~\cite{openai_truncation_tokens}. For simplicity, the token truncation ratio is set to 0.5, implying that when an overflow occurs, the system will discard the earliest half of the tokens.

\subsection{End-to-end Performance}

In the end-to-end experiments, we use 9K conversations from ShareGPT~\cite{sharegpt_sharegptraw_2024} and the average number of turns in these conversations is 5.75. Thus the total number of conversation turns is about 52K. We warm up \storagename using the first 10K conversation turns and then evaluate the performance on the following 42K turns.

\textbf{Cache hit rate.} We first present the cache hit rate of \storagename in CA since other performance metrics are closely related to it.
Figure~\ref{fig:e2e-hit-ratio} shows the total KV cache hit rates, including both DRAM and disk hit rates across various LLMs. 
CA exhibits high hit rates around 86\%, 71\% 89\%, and 90\% for LLaMa-13B, LLaMA-65B, LLaMA-70B, and Falcon-40B, respectively. In contrast, we observe a relatively low hit rate of LLaMA-65B. This discrepancy arises due to the larger storage space required by LLaMA-65B for saving KV caches. Given the same available storage space, CA accommodates fewer sessions for LLaMA-65B, thereby limiting the hit rate. Specifically, LLaMA-65B necessitates 2.5MB of space for each token in the KV cache. LLaMA-13B requires 0.78MB. LLaMA-70B and Falcon-40B require only 0.31MB and 0.12MB of space per token due to using the group query attention with a GQA factor of 8 and 16.

\textbf{Time to first token (TTFT).} TTFT is an important metric for quality of service in LLM serving~\cite{patel2023splitwise, alizadeh2023llm}. It indicates how quickly users start seeing the output of LLMs after entering their prompt. As shown in Figure~\ref{fig:e2e-ttft}, CA significantly reduces the TTFT by 85\%, 61\%, 87\%, and 86\% for LLaMA-13B, LLaMA-65B, LLaMA-70B, and Falcon-40B respectively, in comparison to RE.  
This is because CA eliminates a large amount of repetitive computation for generating the KV caches of historical tokens in the prefilling phase. 
Upon cache hits, the TTFT of CA only relies on the number of newly input tokens in the new conversation turn. 

\textbf{Prefilling throughput.} Prefilling throughput is the metric to evaluate the speed of processing the prompt. Figure~\ref{fig:prefill_throughput} shows the measured prefilling throughput. We observe that CA delivers remarkable speedups of 6.8$\times$, 2.6$\times$, 7.8$\times$ and 7.2$\times$ for LLaMA-13B, LLaMA-65B, LLaMA-70B, and Falcon-40B respectively, when compared to RE. The improvement of CA on prefilling throughput comes from the reduced prefilling time. CA only prefills the new input of the new conversation. Moreover, CA can load and reuse the historical KV caches from \storagename with layer-wise pre-loading optimization. The historical KV cache loading simultaneously occurs with the prefilling on the new input tokens.

\begin{figure}[t]
    \begin{minipage}[t]{0.49\linewidth}
        \includegraphics[width=\linewidth]{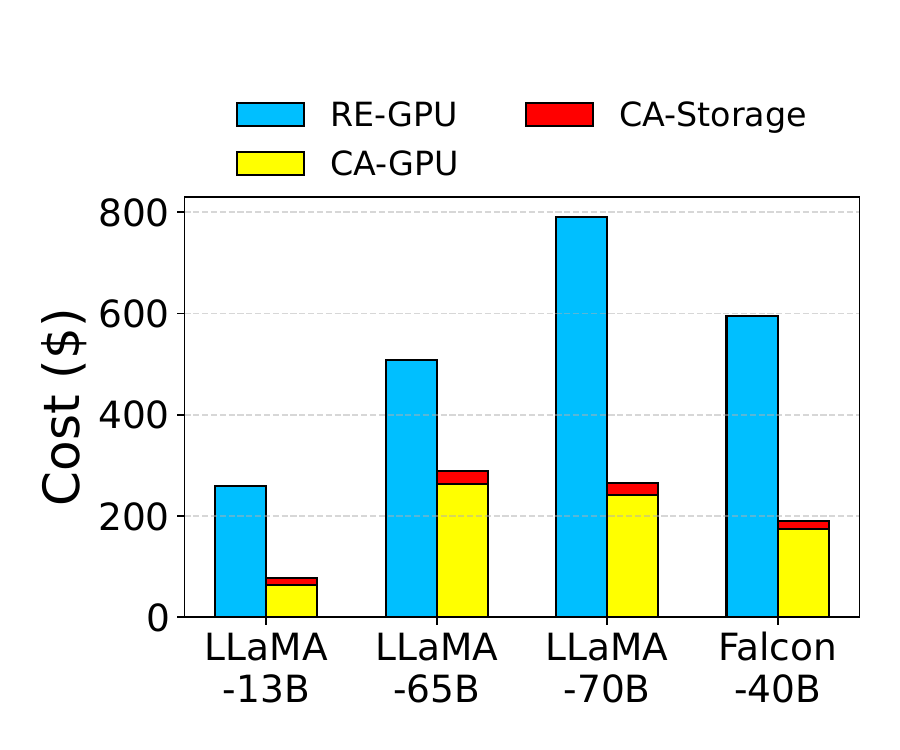}
        \vspace{-20pt}
        \caption{Inference cost.} \label{fig:e2e-cost}
    \end{minipage}
    \hfill
    \begin{minipage}[t]{0.49\linewidth}
      \includegraphics[width=\linewidth]{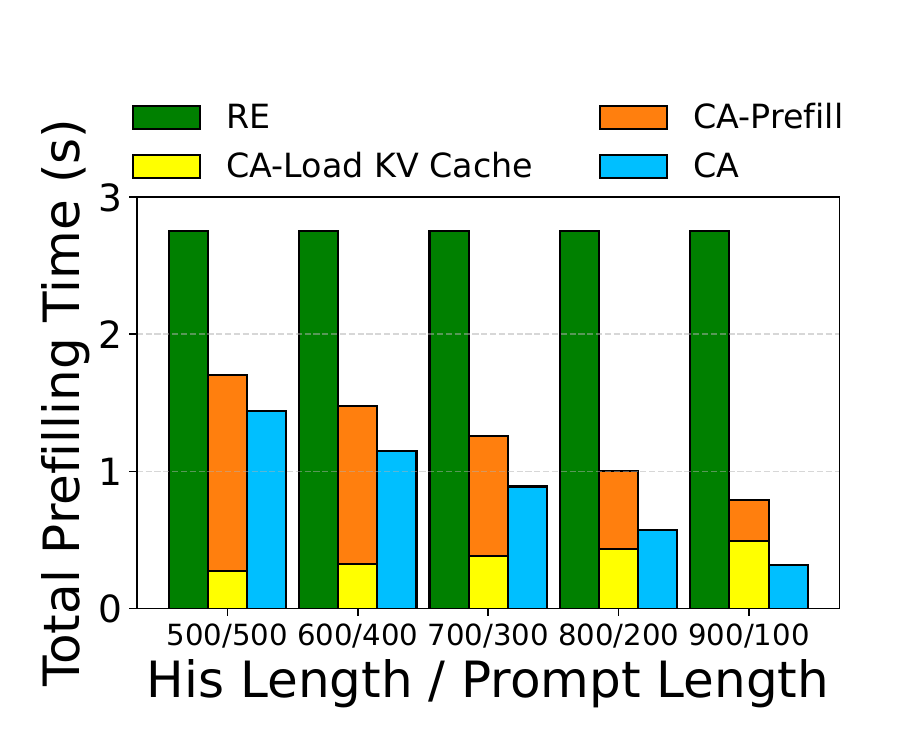}
      \vspace{-20pt}
\caption{Recomputation \textit{v.s.} \attname.} \label{fig:ca-time}
    \end{minipage}
    \vspace{-20pt}
  \end{figure}

\textbf{GPU time.}
Figure~\ref{fig:e2e-performance} shows the end-to-end GPU time to finish all inference jobs in the workload. We observe that CA achieves speedups of 4.0$\times$, 1.9$\times$, 3.3$\times$, and 3.4$\times$ for LLaMA-13B, LLaMA-65B, LLaMA-70B, and Falcon-40B respectively, compared to RE. The performance improvements of CA are from two aspects, which are the mitigation of recomputing KV caches of the historical tokens, and the mitigation of recomputing KV caches after context overflow. 
Regarding the mitigation of re-prefilling, CA efficiently saves the KV cache in \storagename and loads it when necessary for historical tokens. On the other hand, RE discards the KV cache once a job is finished, requiring the redoing of prefilling for every job to reproduce the KV cache.  
In terms of mitigating the recomputation of KV caches after context flow, RE applies token truncation which invalidates the KV cache for each truncation due to the embedded position encoding in the KV caches. This prompts RE to recompute the KV cache based on the truncated historical tokens. In contrast, CA decouples the position information from the KV caches, allowing direct truncation of the KV cache. This approach avoids the recomputation of KV caches that RE requires. Note that \attname also not only aids in minimizing the prefilling time but also helps reduce the decoding time~\cite{yu_orca_2022}. Specifically, under continuous batching, each newly arrived job must complete prefilling before it can join other decoding jobs. This process blocks the execution of decoding jobs, resulting in prolonged decoding time. However, \attname mitigates this issue by minimizing prefilling time for newly arrived jobs. The reduction in prefilling time reflects the decoding time improvement observed in RE, as depicted in Figure~\ref{fig:e2e-performance}.

\textbf{Inference cost.}
We evaluate the resource cost based on the on-demand price of AWS EC2 instances~\cite{ec2_p4d_price_2023, ec2_price_2023}, i.e., \$5/hour per A100 GPU, \$0.0088/hour/GB for DRAM and \$0.000082/hour/GB for SSD.
Figure~\ref{fig:e2e-cost} shows the total costs of different methods for completing the workload. Compared to RE, CA achieves significant cost savings for LLaMA-13B, LLaMA-65B, LLaMA-70B, and Falcon-40B, amounting to 70\%, 43\%, 66\%, and 68\%, respectively. These cost savings primarily stem from the reduced GPU time, as CA effectively reduces redundant prefilling for historical tokens and recomputation costs during context overflow, as depicted in Figure~\ref{fig:e2e-performance}. CA employs cost-effective storage mediums including host memory and disks to cache the KV caches during inactive conversation sessions. The storage cost from the host memory and disks constitutes 16.4\%, 9.0\%, 9.0\%, and 9.0\% of the total cost in CA for LLaMA-13B, LLaMA-65B, LLaMA-70B, and Falcon-40B, respectively. 

\subsection{Ablation Studies}
\label{sec:ablation}

\subsubsection{Recomputation \textit{v.s.} \attname}

We investigate the prefilling performance of different methods under varying historic and new token ratios. Different methods prefill the same 1K tokens under the batch size of 16 on an A100 GPU for LLaMA-13B. RE computes the KV cache for all tokens, while CA loads the KV cache of historical tokens from \storagename and partially prefills the new input tokens. For example, the setting 600/400 means CA loads the KV cache of 600 tokens and computes the KV cache for 400 tokens. As shown in Figure~\ref{fig:ca-time}, CA consistently outperforms RE in all tested settings. This advantage becomes more pronounced as the percentage of newly input tokens decreases (from 500 to 100), as depicted by the middle bar of each bar group. 
Although the KV cache loading time for CA gradually increases with the percentage of historical tokens (from 500 to 900), the layer-wise pre-loading scheme effectively eliminates this loading time, as demonstrated by the third bar of each bar group. 
Note that when the KV cache loading time exceeds the prefilling time (e.g., the second bar of setting 900/100), CA can conceal the KV cache loading time by enabling a read buffer. The impact of the read buffer is evaluated in the next subsection.

\subsubsection{Overlapped KV cache Access}

\begin{figure}[t]
    \begin{minipage}[t]{0.49\linewidth}
        \includegraphics[width=\linewidth]{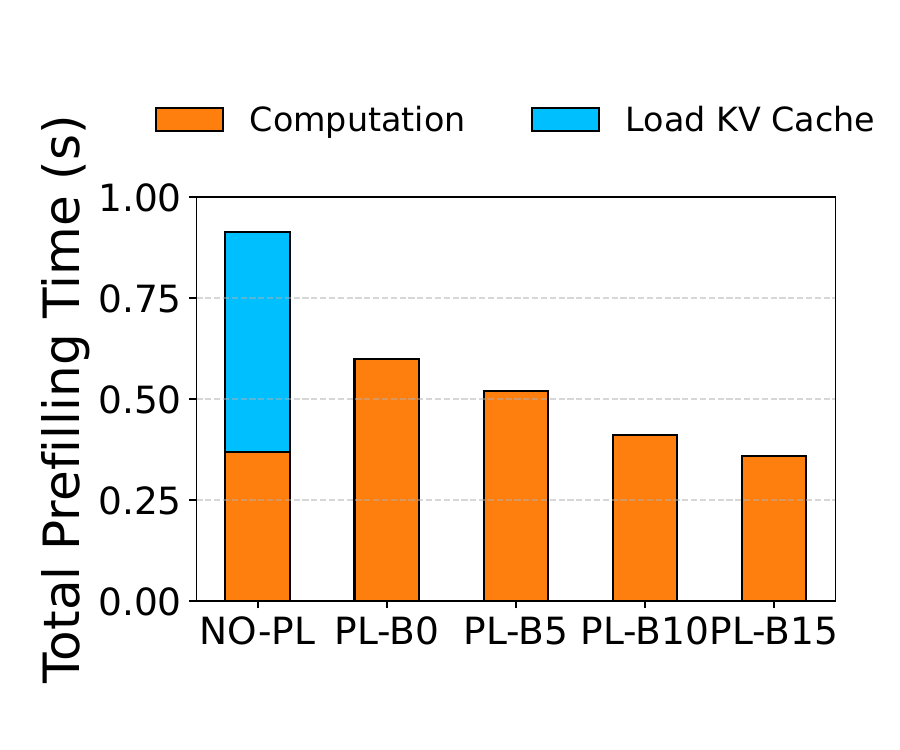}
        \vspace{-25pt}
        \caption{CA with no pre-loading v.s. CA pre-loading with various buffer sizes.} \label{fig:mixoverlap}
    \end{minipage}
    \hfill
    \begin{minipage}[t]{0.49\linewidth}
      \includegraphics[width=\linewidth]{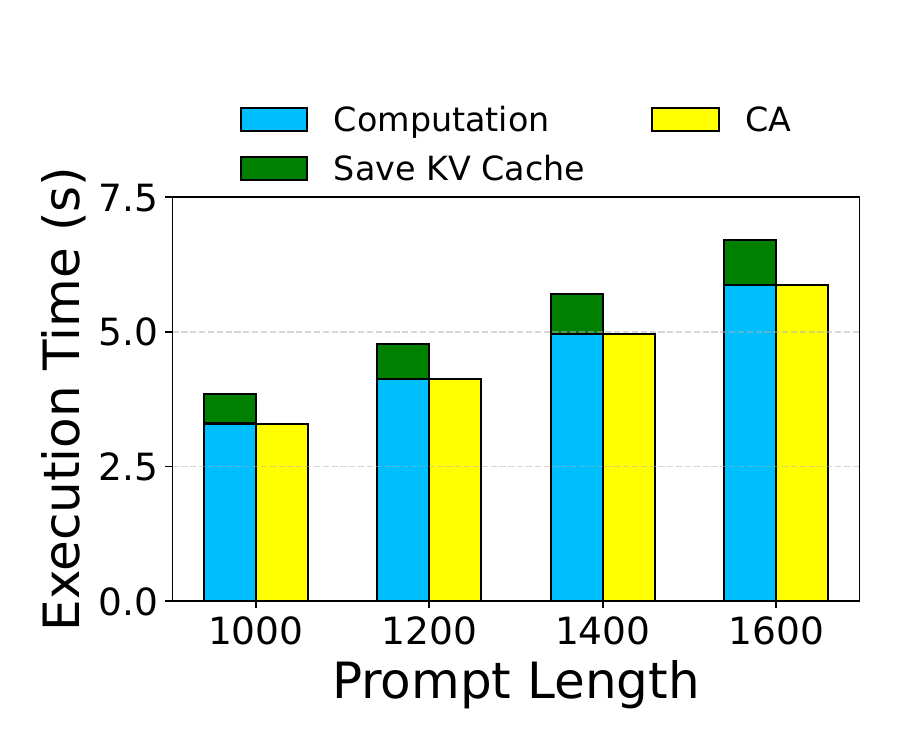}
      \vspace{-25pt}
\caption{Performance impact of using write overlap.} \label{fig:retention}
    \end{minipage}
    \vspace{-10pt}
  \end{figure}

This subsection evaluates the effectiveness of the proposed overlapping access techniques for loading and saving KV caches. The model used is LLaMA-13B with a single GPU and the batch size is set to 16.

\textbf{Layer-wise KV cache pre-loading.} We set the length of historical tokens to 1K and the length of newly input tokens to 100 for investigating the effectiveness of the lay-wise pre-loading scheme. 
The first bar in Figure~\ref{fig:mixoverlap}, i.e., NO-PL, shows the time of prefilling without the pre-loading scheme that includes two parts: the KV cache loading time and the computation time of the newly input tokens.
The following bars in Figure~\ref{fig:mixoverlap} show the prefilling time when the layer-wise pre-loading scheme has different sizes of read buffers. 
For clarity, we use the number of layers to represent the buffer size, e.g., PL-B0 indicates no read buffer and PF-B5 indicates a read buffer size of 5 layers of KV cache. We observe although there is no read buffer, i.e., PL-B0, the pre-loading scheme reduces the prefilling time by 35\% compared to NO-PL. PF-B15 perfectly overlaps the KV cache loading time and reduces the prefilling time by 61\% compared to NO-PL.

\textbf{Asynchronous KV cache saving.} We set different prompt lengths ranging from 1K to 1.6K and the number of decoding steps to 20 for investigating the effectiveness of the asynchronous saving scheme. 
As Figure~\ref{fig:retention} shows, the saving time increases as the prompt length grows, as the size of the KV cache to be saved increases. 
To reduce the saving overhead, \attname employs the asynchronous saving scheme allowing KV cache saving to overlap with inference execution, reducing the overall execution time by 13\% to 15\%.

\subsubsection{Scheduler-aware Fetching and Eviction}
\begin{figure}[t]
    \begin{subfigure}[t]{0.49\linewidth}
        \includegraphics[width=\linewidth]{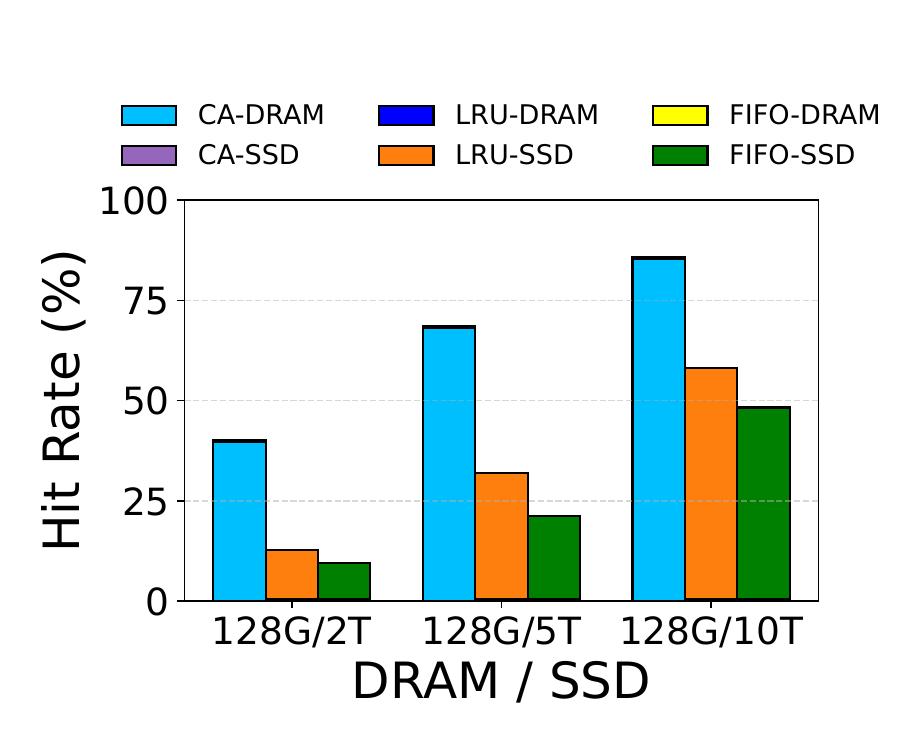}
        \vspace{-20pt}
        \caption{Impact on hit rate.} \label{fig:scheduling-hit-rate}
    \end{subfigure}
    \hfill
    \begin{subfigure}[t]{0.49\linewidth}
      \includegraphics[width=\linewidth]{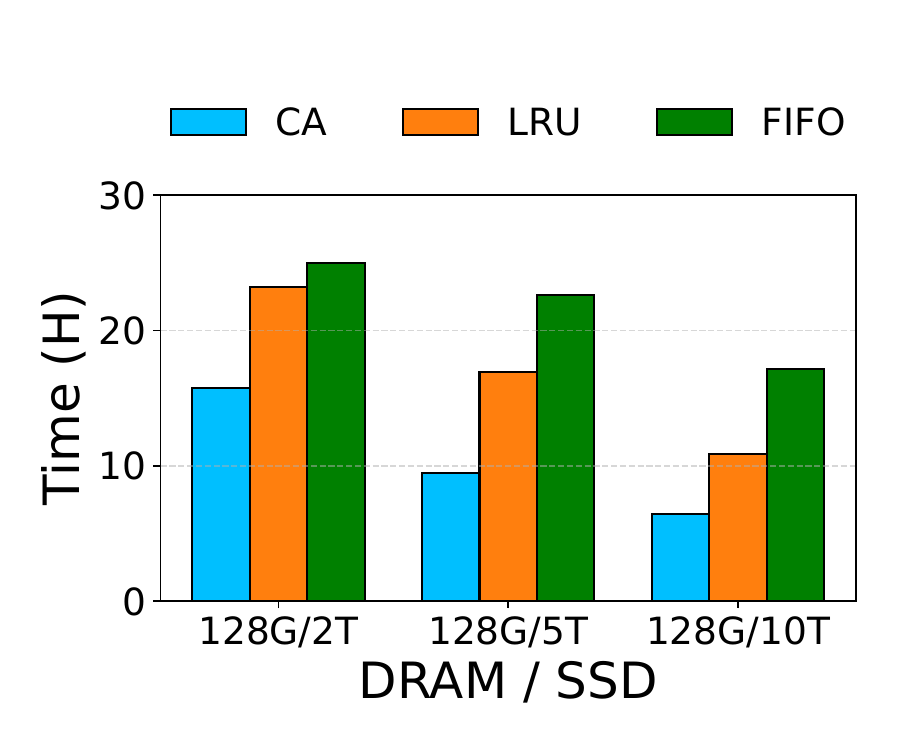}
      \vspace{-20pt}
\caption{Impact on GPU time.} \label{fig:scheduling-throughput}
    \end{subfigure}
    \caption{Comparison of the eviction algorithms under various storage settings.}
    \vspace{-15pt}
    \label{fig:eviction}
  \end{figure}

We evaluate the effectiveness of the scheduler-aware fetching and eviction in \attname upon improving the cache hit rate. We compare overall cache hit rates, DRAM hit rates, and disk hit rates of CA and existing eviction policies (including LRU and FIFO) across various storage configurations.

As shown in Figure~\ref{fig:scheduling-hit-rate}, for the configuration of 128G/2T that indicates 128GB DRAM and 2TB SSD, CA outperforms LRU and FIFO in the overall cache hit rate by 27\% and 31\%, respectively. With the increased SSD capacity (128G/10T), CA achieves a remarkable hit rate of 86\%, surpassing LRU (58\%) and FIFO (48\%). CA achieves high overall hit rates because CA is aware of the future KV cache access information to avoid evicting the KV caches that will be used in the future. The higher hit rates are translated to the reduced GPU time as shown in Figure~\ref{fig:scheduling-throughput}, where CA achieves speedup up to 2.7$\times$.
Analyzing the breakdown of hit rate, for the configuration of 128G/2T, LRU and FIFO only achieve 0.5\% and 0.5\% (too tiny to display) DRAM hit rates, with the remaining 12.4\% and 9.0\% being disk hit rates. Even with the overall hit rate increasing to 58\% and 48\% respectively for the larger capacity of 128G/10T, LRU and FIFO still exhibit limited DRAM hit rates of approximately 0.6\% and 0.5\% respectively. This is because LRU and FIFO lack awareness of future KV cache information and cannot pre-fetch KV caches from disks to host memory, thereby limiting their ability to improve DRAM hit rates. In contrast, CA achieves a cache hit rate of up to 86\%, with over 99.6\% of the hits occurring in DRAM due to its scheduler-aware policy.

\subsubsection{Performance of Decoupled KV Cache Truncation}

\begin{figure}[t]
    \begin{subfigure}[t]{0.49\linewidth}
        \includegraphics[width=\linewidth]{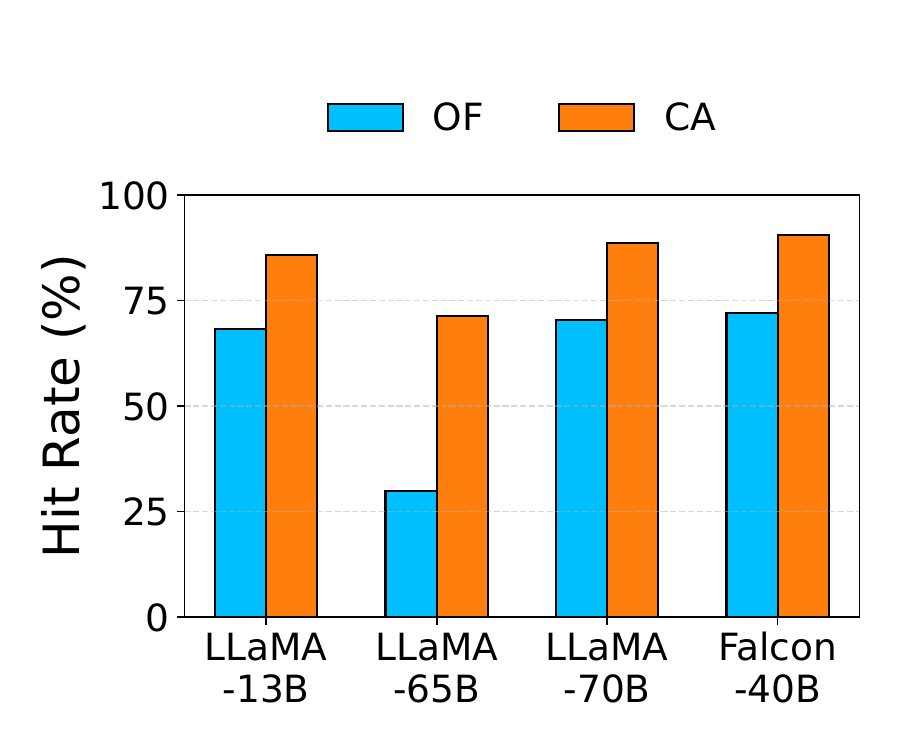}
        \vspace{-20pt}
        \caption{Impact on hit rate.} \label{fig:overflow-hitrate}
    \end{subfigure}
    \hfill
    \begin{subfigure}[t]{0.49\linewidth}
      \includegraphics[width=\linewidth]{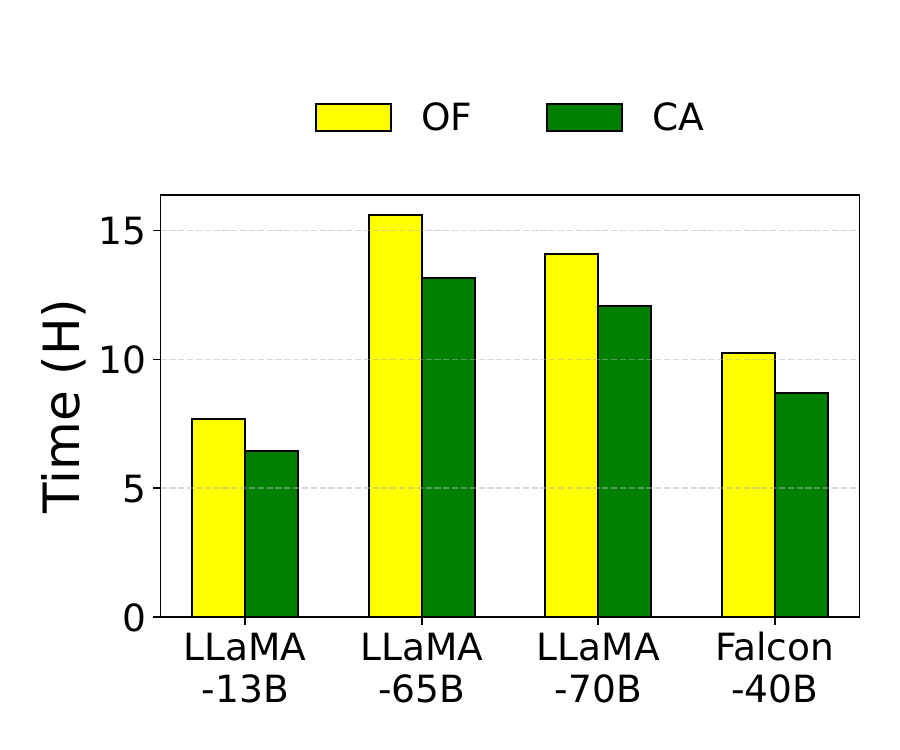}
\vspace{-20pt}
\caption{Impact on GPU time.} \label{fig:overflow-time}
    \end{subfigure}
    \caption{Context overflow impact.}
    \vspace{-15pt}
    \label{fig:eval-overflow}
  \end{figure}

When the context window exceeds its limit, \attname truncates the KV cache directly, thus avoiding the need for re-computation and reducing overhead. We evaluate the effectiveness of the way to manage the context overflow in CA. Specifically, we compare a baseline approach overflow (OF) that embeds positional encoding within the KV caches, leading to the invalidation of KV caches in \storagename. OF relegates context overflow to be managed by recomputation. This experiment uses 128GB DRAM and 10TB SSD. As evident from Figure~\ref{fig:overflow-hitrate}, comparing OF with CA, the hit rates decrease by 17.6\%, 41.5\%, 18.1\%, 18.4\% for LLaMA-13B, LLaMA-65B, LLaMA-70B, and Falcon-40B, respectively. This decline is attributed to the fact that if applying OF, every instance of context overflow necessitates context truncation, thereby invalidating the KV caches in \storagename. This decrease in hit rate subsequently translates to the longer GPU time as shown in Figure~\ref{fig:overflow-time}. 

\attname ensures the validity of the saved KV caches in the system when context overflow happens and promises a higher hit rate and reduced GPU time. OF of LLaMA-65B experiences a low hit rate due to its limited 2K context window.  After serving a conversation of the first turn, the session easily reaches the context window limit, consequently making the associated KV cache invalid. Subsequently, the following conversations in the same session face KV cache miss.

\subsubsection{Accuracy of Decoupled Positional Encoding}
To maintain the validity of the stored KV caches, \attname decouples positional encoding from the KV caches and embeds the new positional encodings when reusing the stored KV caches as presented in Section~\ref{sec:kv-invalidation}. 
We evaluate the impact of the different schemes including CA, token truncation (TT), and naive KV cache truncation (NKVT) on the perplexity (PPL) and the accuracy of LLMs leveraging widely used benchmarks. In situations where the number of historical tokens exceeds the context window limit, TT removes the historical tokens and recomputes the KV caches for the remaining tokens, and NKVT directly discards the KV caches associated with the positional encoding and utilizes the truncated KV caches instead.

\textbf{PPL.} PPL is a metric used to evaluate the quality of a model in generating tokens~\cite{xiao2023efficient, gonen2022demystifying}. Lower PPL values indicate that the model is better at predicting the text and demonstrates a greater understanding of the language. Table~\ref{table:PPL} shows the PPL comparison of LLaMA-7B and LLaMA-13B in CA, TT, and NKVT settings using datasets WikiText-2~\cite{gong2018frage}, C4~\cite{raffel2020exploring}, and PTB~\cite{marcus1993building}. TT consistently achieves a low PPL by recomputing KV caches after context window overflow. CA also maintains a low PPL, comparable to TT (with a difference of $<0.02$), by incorporating new positional encodings into the KV caches after truncation. In contrast, NKVT exhibits a high PPL ($> 10^3$) due to the coupling of positional encoding within its KV caches. Directly truncating the KV caches would scramble the coupled positional information, resulting in the models' failure to maintain a low PPL.

\textbf{Accuracy.}
To analyze the accuracy of the models in answering questions after truncation, we conduct experiments using the MMLU~\cite{hendrycks2020measuring}, LongEval~\cite{li2023long, xiao2023efficient}, and PIQA~\cite{bisk2020piqa} benchmarks. Specifically, we first input a long text to simulate the overflow of historical inputs to trigger the truncation operation, and then append the questions from the benchmarks as new inputs afterward. As shown in Table~\ref{table:ACC}, both CA and TT provide high comparable accuracy. TT achieves high accuracy by paying the recomputation cost for context window overflow, while our CA avoids this cost and still maintains high accuracy. In contrast, the NKVT has a much lower accuracy than CA and TT because the coupled positional encoding after KV cache truncation is miscoded, which results in more disruption to new inputs.

\begin{table}[t]

    \centering
    \caption{PPL comparison of different methods.}
    \label{table:PPL}
\resizebox{0.3\textwidth}{!}{%
    \begin{tabular}{lcccc}
        \toprule
       Dataset & Model & CA & TT & NKVT\\
        \midrule
       \multirow{2}{*}{WikiText-2} & LLaMA-7B & 5.47 &  5.48 & 2198.7 \\
       & LLaMA-13B & 4.91 & 4.90 & 1647.7 \\
       \midrule
       \multirow{2}{*}{PTB} & LLaMA-7B & 8.48 &  8.49 & 2543.5\\
       & LLaMA-13B & 7.61 & 7.60 & 1865.8 \\
       \midrule
       \multirow{2}{*}{C4} & LLaMA-7B & 6.96 &  6.98 & 2343.5\\
       & LLaMA-13B & 6.44 & 6.45 & 1745.6 \\
        \bottomrule
    \end{tabular}
}
\end{table}

\begin{table}[t]
    \centering
    \caption{Accuracy of different methods.}
    \label{table:ACC}
\resizebox{0.3\textwidth}{!}{%
    \begin{tabular}{lcccc}
        \toprule
       Benchmark & Model & CA & TT & NKVT\\
        \midrule
        \multirow{2}{*}{MMLU} & LLaMA-7B & 43.7\% & 43.4\% & 21.8\% \\
         & LLaMA-13B & 52.3\% & 53.2\% & 29.6\% \\
         \midrule
        \multirow{2}{*}{LongEval} & LLaMA-7B & 66.0\% & 65.9\% & 12.0\% \\
         & LLaMA-13B & 68.0\% & 68.0\% & 14.0\% \\
         \midrule
        \multirow{2}{*}{PIQA} & LLaMA-7B & 77.1\% & 77.2\% & 48.9\% \\
         & LLaMA-13B & 80.5\% & 80.4\% & 50.2\% \\
           
    \bottomrule
    \end{tabular}
}
\end{table}

\subsubsection{The Cache Capacity Requirement} \label{sec:eval-capacity}

\begin{figure}[t]
    \begin{subfigure}[t]{0.49\linewidth}
        \includegraphics[width=\linewidth]{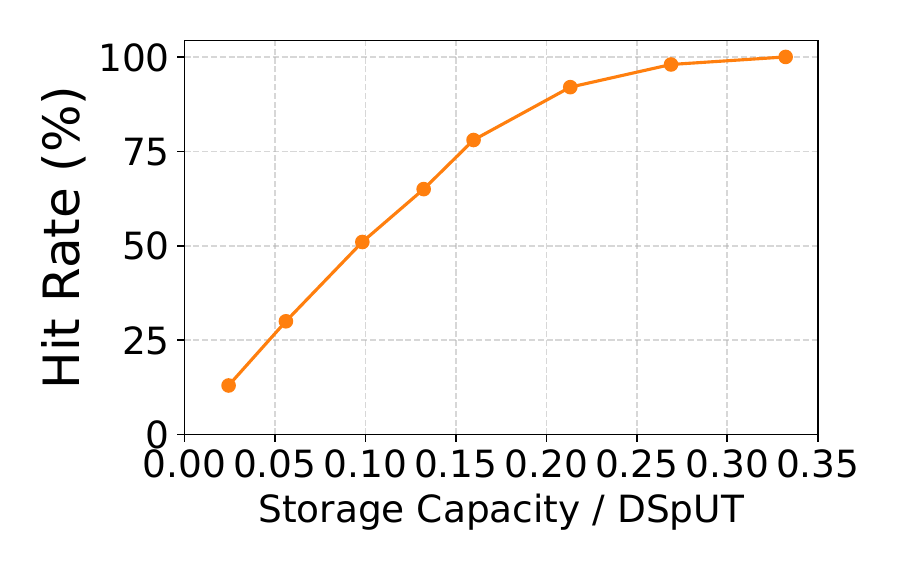}
        \vspace{-20pt}
        \caption{Impact on hit rate.} \label{fig:capacity-hit-rate}
    \end{subfigure}
    \hfill
    \begin{subfigure}[t]{0.49\linewidth}
      \includegraphics[width=\linewidth]{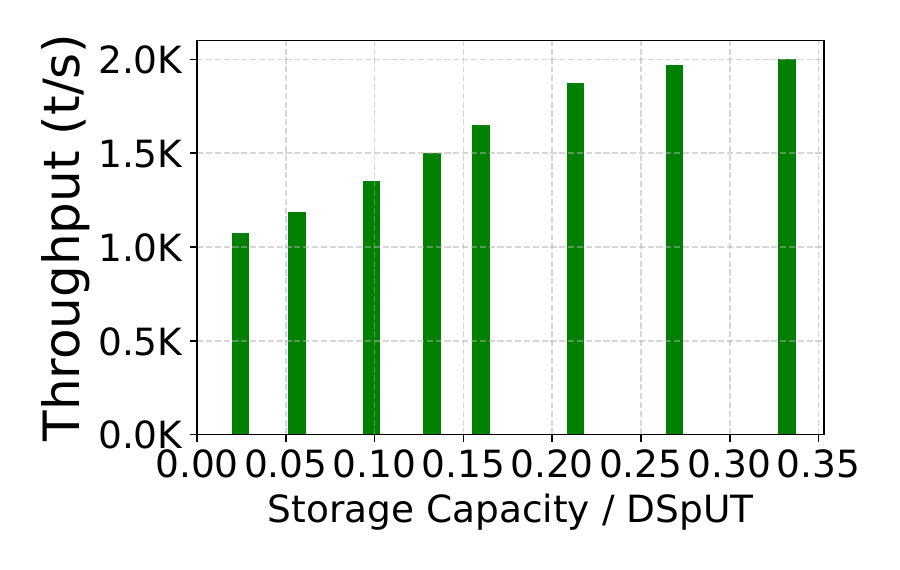}
      \vspace{-20pt}
\caption{Impact on throughput.} \label{fig:capacity-tp}
    \end{subfigure}
    \caption{Impact of storage capacity and the number of distinct sessions.}
    \vspace{-10pt}
  \end{figure}

In this subsection, we investigate how much cache capacity \attname needs to achieve a remarkable cache hit rate. The cache capacity required is related to the maximum number of distinct conversation sessions served by an LLM serving system per unit time (denoted as $\textit{DSpUT}$). The larger the $\textit{DSpUT}$ value is, the more distinct sessions the system handles per unit time, and the more KV cache storage space is required. Moreover, due to the limitation of the maximum context window, the maximum KV cache capacity required by one conservation session is fixed, i.e., equal to the length of the maximum context window multiplied by the KV size of each token, which is denoted as $\textit{CCpS}$. Thus the required maximum cache capacity per unit time is $\textit{CCpUT} = \textit{DSpUT} * \textit{CCpS}$. In \attname, the KV cache of each session has a TTL (time to live) that indicates its maximum saving time since the last access. The TTL is set as the unit time mentioned above. By configuring the cache capacity of \attname as $\textit{CCpUT}$, we can achieve a cache hit rate of 100\% if not considering the newly arrived conversations.
Nevertheless, to achieve a high cache hit rate in real scenarios, we do not need to configure such a large capacity since the hotness of cached items is different.

To figure out the relationship between the required cache capacity ($RCC$) and $CCpUT$, we evaluate the cache hit rate and the decoding throughput under different ratios of  $RCC$ to $CCpUT$. In the experiment, we set the TTL to one hour. As shown in Figure~\ref{fig:capacity-hit-rate}, when the ratio $RCC / CCpUT$ is 0.1, we achieve the cache hit rate of 51\%. When the ratio $RCC / CCpUT$ is 0.25, we achieve the cache hit rate of 98\%. As the hit rate reaches the peak, the throughput also meets its peak, as shown in Figure~\ref{fig:capacity-tp}.

\subsubsection{Impact of Caching Storage Mediums}
\label{sec:diff-storage-medium}

\begin{figure}[t]
    \begin{subfigure}[t]{0.49\linewidth}
        \includegraphics[width=\linewidth]{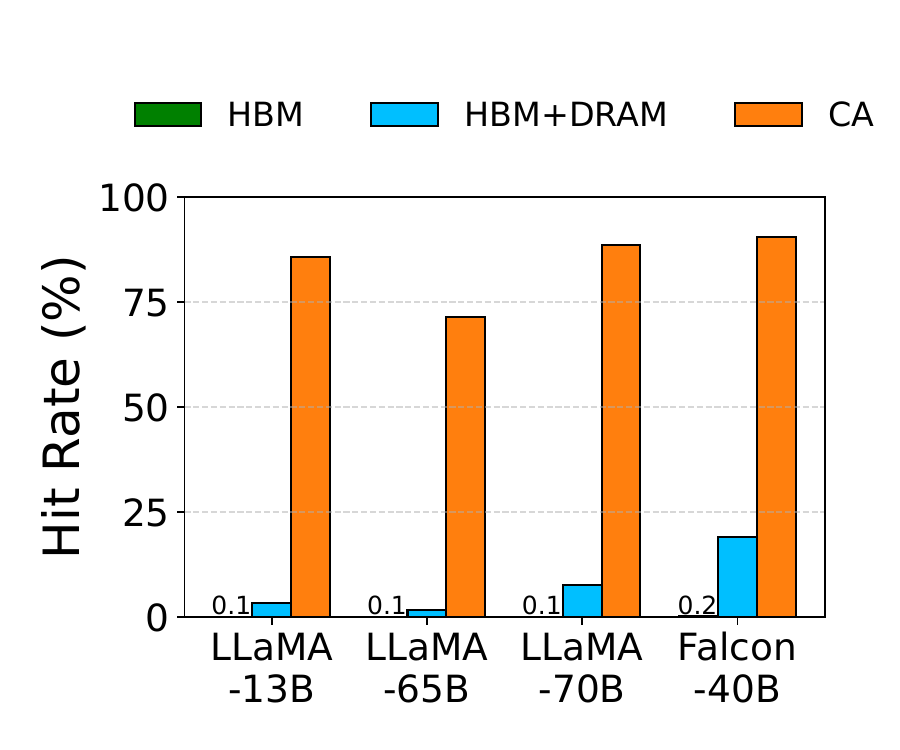}
        \vspace{-20pt}
        \caption{Impact of caching storage mediums on hit rates.} \label{fig:hbm-hit-rate}
    \end{subfigure}
    \hfill
    \begin{subfigure}[t]{0.49\linewidth}
      \includegraphics[width=\linewidth]{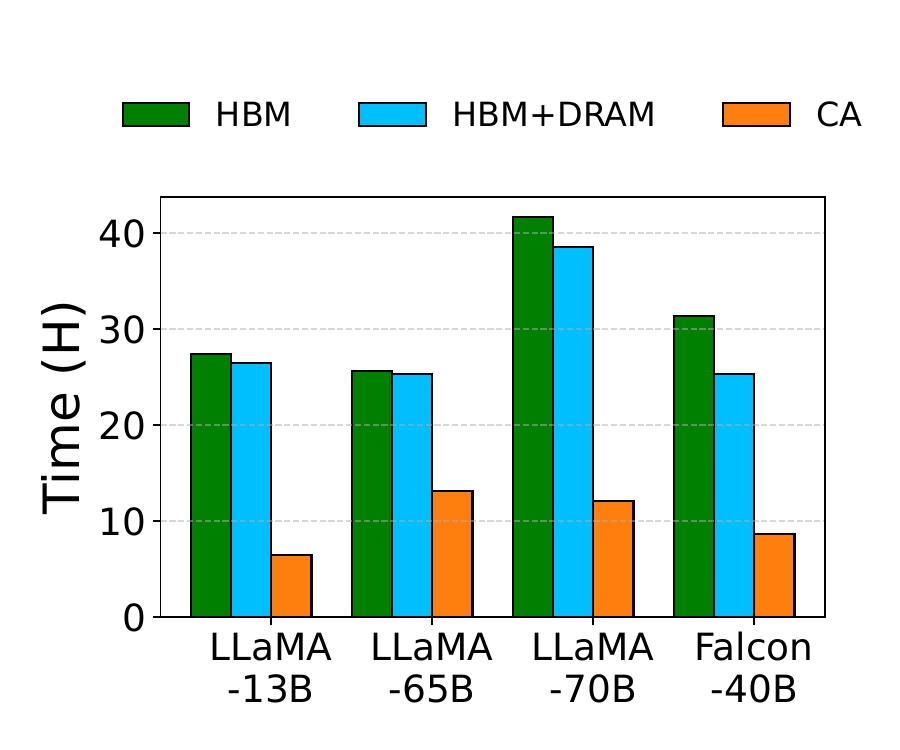}
    \vspace{-20pt}
\caption{Impact of caching storage mediums on GPU time.} \label{fig:hbm-time}
    \end{subfigure}
    \caption{Performance under various caching configurations.}
    \vspace{-10pt}
    \label{fig:with_hbm}
  \end{figure}

Some existing works \cite{zheng2023efficiently,lmdeploy_2023, nvidia_fastertransformer_2023} employ only the HBM space for caching the KV caches of multi-turn conversations. We here compare the performance of mechanism caching KVs on HBMs with that of \attname caching KVs on DRAM and SSDs. In the experiments, we configure the size of the HBM cache as 10GB, the size of DRAM as 128GB, and the size of SSDs as 10TB. Figure~\ref{fig:hbm-hit-rate} shows cache hit rates and inference performance of different mechanisms. The hit rate of the HBM-only caching method is nearly 0\% for all models due to the limited capacity of HBM. 
Using HBM with DRAM improves the cache hit rate to 3.4\%, 1.7\%, 7.7\%, and 19.1\% for models LLaMA-13B, LLaMA-65B, LLaMA-70B and Falcon-40B, respectively. In contrast, by further extending the cache capacity with SSDs, \attname improves the cache hit rate to 86\%, 71\%, 89\%, and 90\% for models LLaMA-13B, LLaMA-65B, LLaMA-70B, and Falcon-40B, respectively. With higher hit rates, \attname significantly improves the inference performance compared to the HBM-only/HBM+DRAM policies as shown in  Figure~\ref{fig:hbm-time}.

\begin{figure}[t]
     \begin{subfigure}[h]{0.24\linewidth}
         \includegraphics[width=\linewidth]{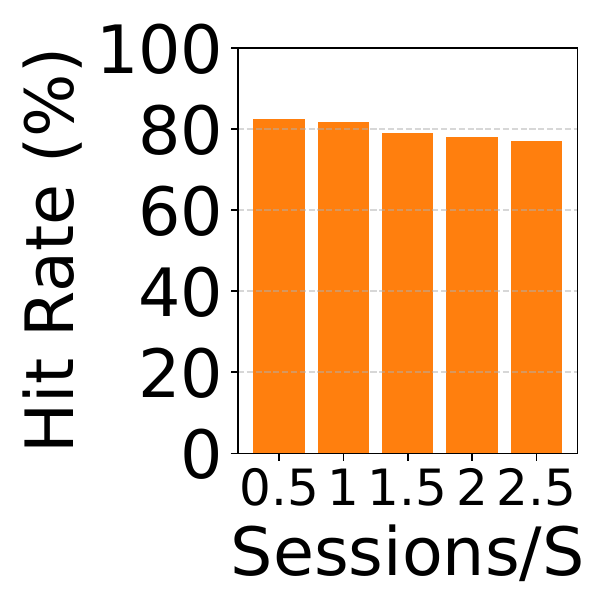}
        \caption{Hit rates.} 
        \label{fig:exp:arrival-rates:hit-rates}
     \end{subfigure}
     \begin{subfigure}[h]{0.24\linewidth}
         \includegraphics[width=\linewidth]{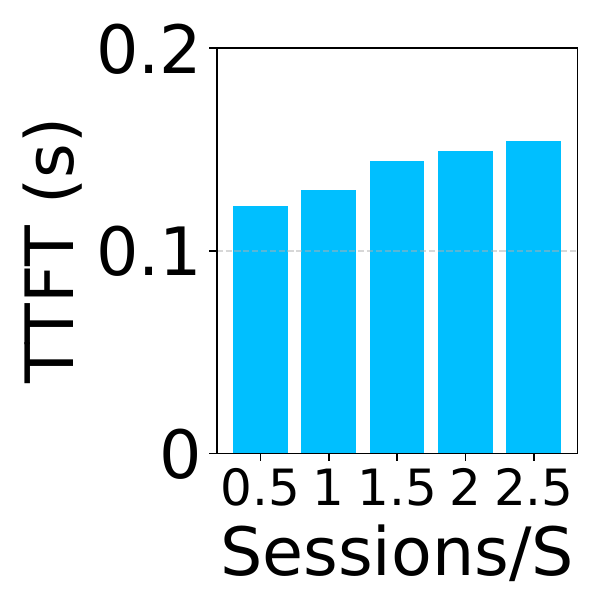}
        \caption{TTFT.} 
        \label{fig:exp:arrival-rates:TTFT}
     \end{subfigure}
    \begin{subfigure}[h]{0.24\linewidth}
         \includegraphics[width=\linewidth]{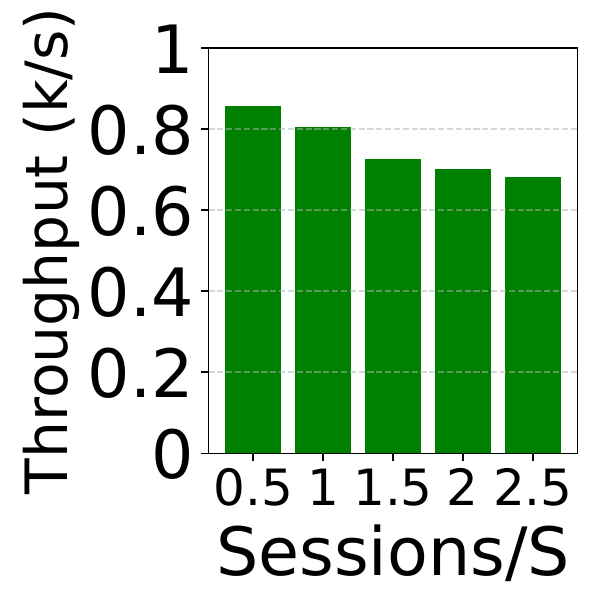}
        \caption{Prefill TP.} 
        \label{fig:exp:arrival-rates:prefill-TP}
     \end{subfigure}
    \begin{subfigure}[h]{0.24\linewidth}
         \includegraphics[width=\linewidth]{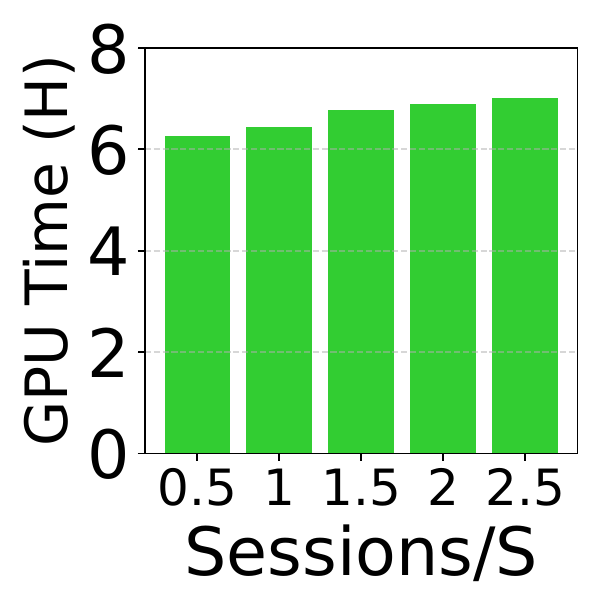}
        \caption{GPU time.} 
        \label{fig:exp:arrival-rates:gpu-time}
     \end{subfigure}
     \caption{Impact of session arrival rates.} 
     \vspace{-15pt}
     \label{fig:exp:arrival-rates}
\end{figure}

\subsubsection{Impact of Session Arrival Rates}

We evaluate how the session arrival rates affect the performance of the \attname in terms of hit rates, TTFT, prefilling throughput, and GPU time. In the experiments, we configure the session arrival rates varying from 0.5/s to 2.0/s. We test LLaMA-13B with 128GB/10TB storage following warmup as before. As shown in Figure~\ref{fig:exp:arrival-rates:hit-rates}, the cache hit rate has a slight drop down, from 82\% to 77\%, as the session arrival rates increase from 0.5 to 2. 
If the session arrival rates are low, a specific time window will contain fewer sessions, indicating a smaller \textit{DSpUT} discussed in Section~\ref{sec:eval-capacity}. The \textit{DSpUT} at low session arrival rates is less than that at high session arrival rates. For equivalent hit rates, a lower \textit{DSpUT} necessitates less cache storage space. In other words, given the same cache storage, a lower \textit{DSpUT} results in higher hit rates, as the results observed in Figure~\ref{fig:exp:arrival-rates:hit-rates}.
Consequently, due to the KV cache being less hit, more sessions need to recompute the KV cache, leading to an increase in the TTFT from 0.122s to 0.154s, as shown in  Figure~\ref{fig:exp:arrival-rates:TTFT}. The prefilling throughput shown in Figure~\ref{fig:exp:arrival-rates:prefill-TP} also experiences a decline, dropping from 858K/s to 681K/s, as an outcome of the slightly decreased KV cache hits. The GPU time shown in Figure~\ref{fig:exp:arrival-rates:gpu-time} escalates from 6.25H to 7.01H. Overall, the impact of varying session arrival rates has a minimal impact on the effectiveness. Note that even with an increase in session arrival rates, the KV cache hit rates remain significant, implying that \attname can still effectively leverage the KV cache, consistently offering substantial performance improvement over recomputation.

%% file: tex/related.tex
\section{Related Work}
\label{sec:related}
\textbf{KV Cache Management.}
Within a single-turn conversation, the KV cache is widely used for improving the performance of the decoding phase~\cite{llm_vllm_2023,yu_orca_2022, zheng2023efficiently, del2023skipdecode, spector2023accelerating, wang2023tabi}. To reduce the storage overhead of the KV cache on HBMs, existing work employs quantization and compression techniques on KV caches~\cite{xiao2023efficient, ge2023model, zhang2023h, liu2023scissorhands}. To reduce the memory waste incurred by fragmentation, vLLM \cite{kwon_efficient_2023} takes inspiration from virtual memory to allow the KV cache to use fine-granularity non-continuous memory. These techniques are orthogonal to \attname, which focuses on multi-turn conversations. 

LMDeploy~\cite{lmdeploy_2023} is an LLM inference framework that caches the KV caches of multi-turn conversations on HBMs.
RadixAttention \cite{zheng2023efficiently}, ChunkAttention~\cite{ye2024chunkattention}, and  Pensieve~\cite{yu2023stateful} are inference techniques that were developed concurrently with \attname. RadixAttention and ChunkAttention optimize the inference tasks that share prompt prefixes. Tasks with the same prompt prefixes share the same KV caches to reduce the KV computation. Pensieve utilizes both GPU and CPU memory to store KV caches for multi-turn conversations. 
Different from all these works, \attname exploits slower but larger storage hierarchies to save the KV caches to achieve high cache hit rates as presented in Section~\ref{sec:diff-storage-medium}, and focuses on designing systemic techniques to address the challenges of offloading to slower mediums.

\textbf{Inference Parameter Offloading.}
FlexGen \cite{sheng_glexgen_2023} offloads both model weights and KV cache to DRAM and disks to support offline inference of LLMs. DeepSpeed Inference \cite{aminabadi_deepspeed_2022, rajbhandari_zero_2021, rajbhandari2022deepspeed} offloads model weights to the DRAM and disks and fetches them on demand. Lina \cite{li_accelerating_2023} offloads infrequently used expert weights of LLMs to the host memory to improve memory efficiency. PowerInfer \cite{song2023powerinfer} and LLM in a flash\cite{alizadeh2023llm} utilize sparsity \cite{liu2023deja, mirzadeh2023relu} in FFN computation to offload most of the inactive weights to the host memory or disks to reduce both memory usage and the computation.  
FastServe \cite{wu_fast_2023} schedules the KV caches to the host memory for optimizing the job completion time. 
In contrast, \attname exploits KV cache offloading to reduce the recomputation overhead in multi-turn conversations. 

%% file: tex/conclusion.tex
\section{Conclusion}
\label{sec:conc}
This paper proposes \attname, a new attention that reuses the KV caches for any ensuing turns of the same conversation, achieving a significant reduction in the recomputation overhead of KV caches in LLMs. To improve the efficiency of \attname, we design overlapped KV cache access, hierarchical KV cache
placement, and positional encoding decoupled KV cache truncation schemes. Extensive experimental results demonstrate that \attname significantly decreases the TTFT by up to 87\% and improves the prompt prefilling throughput by 7.8$\times$ for multi-turn conversations. It reduces the end-to-end inference cost by up to 70\%. 